\documentclass[
]{ceurart}

\usepackage{tabularx}
\usepackage{multirow}
\usepackage{tikz}
\usepackage{booktabs}
\usepackage{listings}
\usepackage{todonotes}
\usepackage{graphicx}

\usetikzlibrary{backgrounds, fit, positioning, arrows.meta}
\begin{document}

\copyrightyear{2026}
\copyrightclause{Copyright for this paper by its authors.
  Use permitted under Creative Commons License Attribution 4.0
International (CC BY 4.0).}

\conference{CLEF 2026 Working Notes, 21 -- 24 September 2026, Jena, Germany}

\title{BIT.UA at BioASQ 14B: Modular Retrieval with pg\_textsearch and Qdrant, and Agent-Based Answer Generation}

\title[mode=sub]{BIT.UA at CLEF 2026}

\author[2]{André Ribeiro}[%
  orcid=0009-0008-0194-7171,
  email=andrepedroribeiro@ua.pt,
]
\cormark[1]

\author[1]{Rúben Garrido}[%
  orcid=0009-0002-1508-7885,
  email=rubengarrido@ua.pt,
]
\cormark[1]

\author[3]{Alexander Christiansen}[%
  orcid=0009-0003-3867-7324,
  email=ach22@student.aau.dk,
]

\author[1,3]{Richard A. A. Jonker}[%
  orcid=0000-0002-3806-6940,
  email=richard.jonker@ua.pt,
]


\author[1]{Sérgio Matos}[%
  orcid=0000-0003-1941-3983,
  email=aleixomatos@ua.pt,
]

\address[1]{IEETA/DETI, LASI, University of Aveiro, Aveiro, Portugal}
\address[2]{IT, University of Aveiro, Aveiro, Portugal}
\address[3]{Aalborg University Business School, Aalborg, Denmark}

\cortext[1]{Corresponding author.}

\begin{abstract}
  This paper describes the participation of the BIT.UA team from the University of Aveiro in the 14th edition of the BioASQ Task B challenge on biomedical question answering. Building on our previous submissions, we introduced a substantially refactored and modular codebase, and made significant changes to both the retrieval and generation components of the pipeline. For Phase~A document retrieval, we replaced the PyTerrier PISA index with PostgreSQL-based pg\_textsearch for BM25 retrieval and adopted Qdrant for dense embedding indexing, enabling more efficient storage and GPU-accelerated similarity search. We explored HyDE-based query expansion alongside a Context-1 retrieval strategy. A new reranker training pipeline was developed, incorporating dense retrieval for negative sampling. For Phases A+ and B answer generation, we introduced an LLM-as-a-judge framework and a novel agent quorum mechanism, where multiple agents with diverse prompts debate and iteratively converge on a consensus answer using adaptive document retention. We also participated in the snippets generation subtask for the first time. Our systems achieved competitive results across all batches, with Phase~A systems achieving MAP ranks of 5 (Batch~1,3). We discuss the impact of these architectural changes, lessons learned, and outline directions for future work including SPLADE and ColBERT integration. All code is openly available: \url{https://github.com/bioinformatics-ua/BioASQ14b}.
\end{abstract}


\maketitle
\section{Introduction}
\label{sec:introduction}

Biomedical question answering remains a critical and challenging task in natural language processing, driven by the accelerating growth of biomedical literature. Systems capable of retrieving relevant documents and generating accurate, evidence-based answers to complex biomedical questions are essential for researchers and clinicians navigating an ever-expanding knowledge base. The BioASQ Task~B challenge~\cite{BioASQ2026,QAcorpusBioASQ} has consistently provided a rigorous benchmark for evaluating and advancing these technologies, requiring participants to retrieve relevant PubMed documents (Phase~A) and generate answers from them (Phases~A+~and~B).

Our team, BIT.UA from the University of Aveiro, has participated in several consecutive editions of the BioASQ challenge~\cite{almeida2024bit, jonker2025bit}. Each year, we have iteratively refined our pipeline based on lessons learned from both automatic metrics and, more critically, human evaluation results. The 13th edition~\cite{jonker2025bit} revealed a persistent misalignment between automatic evaluation metrics and human-judged answer quality, motivating a fundamental rethinking of our approach. In that edition, we also observed that Dense Pseudo Relevance Feedback (DPRF) yielded consistent improvements in retrieval, while our generation systems remained our weakest component.

For the 14th edition, we undertook a major architectural overhaul, with new reranker models. The entire codebase was restructured with a focus on modularity, enabling cleaner separation of concerns and more flexible experimentation. In Phase~A, we replaced the PyTerrier PISA indexing system with PostgreSQL-based pg\_textsearch for BM25 retrieval, integrated Qdrant for dense embedding indexing, adopted the Text Embeddings Interface (TEI) for efficient embedding generation, and developed a new reranker training pipeline. We also explored query expansion via HyDE and a Context-1 LLM-based retrieval strategy. For Phases~A+~and~B, we introduced an LLM-as-a-judge framework and a novel multi-agent quorum mechanism, where agents with diverse prompts debate and converge on a consensus answer using adaptive document retention based on snippet-level evidence. We also extended our participation to the snippet generation subtask for the first time, and revisiting exact answer generation.

This paper is organized as follows. Section~\ref{sec:previous-work} summarizes our system from the previous year and the conclusions that shaped this year's development. Section~\ref{sec:methodology} describes the methodological changes introduced across all pipeline phases, Section~\ref{sec:results} presents our internal validation and official results, Section~\ref{sec:discussion} discusses the lessons learned and directions for future work, and Section~\ref{sec:conclusion} concludes the paper.

\section{Previous Work}
\label{sec:previous-work}

In our participation in BioASQ~13B~\cite{jonker2025bit}, our system comprised a hybrid two-stage retrieval pipeline for Phase~A and a retrieval-augmented generation (RAG) framework for Phases~A+~and~B.

\subsection{Phase A: Document Retrieval}

For document retrieval, we employed BM25-based first-stage retrieval using the PISA framework~\cite{mallia2019pisa, macavaney2022python} to efficiently filter candidate documents from the PubMed corpus. This was followed by neural reranking using transformer-based cross-encoders, specifically PubMedBERT~\cite{gu2021domain} and BioLinkBERT~\cite{yasunaga2022linkbert}, applied to the top 1,000 BM25-retrieved documents. To further enhance retrieval, we incorporated Dense Pseudo Relevance Feedback (DPRF) using BGE-M3~\cite{chen2024bge} embeddings. The outputs from multiple rerankers were combined using reciprocal rank fusion (RRF).

Key conclusions from Phase~A in the 13th edition included: (1)~base model variants generally performed comparably to their large counterparts; (2)~higher data quality led to better performance than data quantity, a finding confirmed by official evaluation; (3)~DPRF yielded consistent although modest improvements; and (4)~reducing the number of reranked documents from 1,000 to 100 had negligible impact on performance while enabling much faster inference. Our Phase~A systems consistently achieved top rankings, with DPRF-based submissions outperforming larger ensembles in most batches.

\subsection{Phases A+ and B: Answer Generation}

For answer generation, we adopted a RAG framework providing the top retrieved documents as context to large language models including Llama~3~70B, Nous-Hermes2-Mixtral, LLaMA~Nemotron~70B, OpenBioLLM~\cite{pal2024openbiollms}, and a custom fine-tuned Gemma~3~27B model~\cite{gemma2025}. The Gemma model was fine-tuned in two stages using LoRA~\cite{hu2022lora}: first on a custom biomedical dataset for knowledge injection, then specifically for RAG tasks.

A key architectural change was the unification of ensembling and summarization into a single generation step, where a summarization prompt combined outputs from multiple models into one coherent answer. We also transitioned our inference engine from Ollama to LMDeploy~\cite{lmdeploy2023} for faster generation.

Critical insights from the 13th edition included the persistent misalignment between automatic evaluation metrics and human judgments. Our system which achieved a high ROUGE-based F1 scores was rated poorly by human evaluators, and vice versa. This discrepancy exposed fundamental limitations in relying on automatic metrics alone and motivated a shift towards prioritizing robustness and consistency across batches. The lack of intermediate results between batches in the 13th edition inadvertently encouraged us to focus on cross-batch generalization rather than short-term metric optimization.

Our generation systems remained the weakest component across editions. While summarization techniques yielded modest F1 gains at the expense of recall, these changes translated into substantial improvements in human evaluation scores. The use of snippets consistently produced top recall scores, though this did not always translate into high human evaluation rankings.

These findings directly informed our 14th edition strategy, where we sought to address the architectural and methodological limitations identified in our previous work while maintaining the elements that proved effective.

\section{Methodology}
\label{sec:methodology}

This section describes the major architectural and methodological changes introduced in the 14th edition of our BioASQ participation. Beyond the specific pipeline modifications described below, we undertook a comprehensive codebase refactoring, restructuring the entire project with a focus on modularity. This refactoring separated retrieval, reranking, and generation components into distinct, independently testable modules, facilitating more rapid experimentation and reducing coupling between pipeline stages.

\subsection{Architecture Overview}

\begin{figure}[ht]
  \centering
  \resizebox{\linewidth}{!}{%
  \begin{tikzpicture}[
      node distance=0.6cm and 0.6cm,
      box/.style={draw, rounded corners=3pt, minimum width=2.0cm, minimum height=0.7cm, align=center, font=\scriptsize},
      widebox/.style={draw, rounded corners=3pt, minimum width=2.4cm, minimum height=0.7cm, align=center, font=\scriptsize},
      fusionbox/.style={draw, rounded corners=3pt, minimum width=1.6cm, minimum height=0.7cm, align=center, font=\scriptsize},
      smallbox/.style={draw, rounded corners=2pt, minimum width=1.2cm, minimum height=0.5cm, align=center, font=\tiny},
      altbox/.style={draw, dashed, rounded corners=5pt, minimum width=8.2cm, minimum height=1.6cm, fill=yellow!6},
      arr/.style={-{Stealth[scale=0.7]}, thick},
      dasharr/.style={-{Stealth[scale=0.7]}, thick, densely dashed},
    ]

    \node[box, fill=blue!10] (bm25) {BM25\\(pg\_textsearch)};
    \node[box, fill=blue!10, below=0.3cm of bm25] (dense) {Dense\\(BGE-M3 + Qdrant)};
    \node[smallbox, fill=yellow!15, left=0.4cm of dense] (hyde) {HyDE};
    \draw[dasharr] (hyde.east) -- (dense.west);

    \node[fusionbox, fill=green!10, right=0.8cm of bm25, yshift=-0.5cm] (fusion) {Fusion\\RRF / wsum};
    \draw[arr] (bm25.east) -- ++(0.3,0) |- (fusion.west);
    \draw[arr] (dense.east) -- ++(0.3,0) |- (fusion.west);

    \node[widebox, fill=orange!12, right=0.6cm of fusion] (rerank) {Re-ranking\\Cross-encoders};
    \node[widebox, fill=red!10, right=0.6cm of rerank] (snippets) {Snippets\\Gemma~4};
    \node[widebox, fill=purple!10, right=0.6cm of snippets] (gen) {Generation\\Agents / LLM judge};

    \draw[arr] (fusion.east) -- (rerank.west);
    \draw[arr] (rerank.east) -- (snippets.west);
    \draw[arr] (snippets.east) -- (gen.west);

    \node[altbox, below=2.0cm of rerank.south] (ctx1-outer) {};
    
    \node[font=\scriptsize\bfseries, above=0.3cm of ctx1-outer.north, anchor=east] {Context-1 (Agentic Retrieval)};

    \node[smallbox, fill=blue!10] at ([xshift=-1.6cm, yshift=-0.1cm]ctx1-outer.center) (ctxdense) {Dense};
    \node[smallbox, fill=blue!10, left=0.3cm of ctxdense] (ctxbm25) {BM25};
    \node[font=\tiny, above=0.05cm of ctxbm25.north, anchor=south, xshift=0.75cm] {tools};
    \node[font=\scriptsize, right=0.8cm of ctxdense, text width=4.0cm, align=center] {Multi-turn agentic\\search with self-\\editing context};

    \draw[dasharr] (ctx1-outer.north) -| (rerank.south);

    \node[font=\scriptsize\bfseries, text=blue!60, above=0.3cm of bm25.north west, anchor=south west] {Phase A — Traditional Pipeline};

    \node[font=\scriptsize\bfseries, text=blue!60, above=0.4cm of ctx1-outer.north, anchor=south east] {Phase A — Agentic Alternative};
    
    \node[font=\scriptsize\bfseries, text=purple!60, above=0.3cm of gen.north west, anchor=south west] {Phases A+ / B};
    
  \end{tikzpicture}%
  }
  \caption{Pipeline architecture for the 14th edition. The traditional Phase~A path (top) fuses BM25 and dense retrieval, re-ranks, and extracts snippets. Context-1 (bottom) is a fully independent agentic pipeline that uses BM25 and dense search as internal tools, producing a ranked document set.}
  \label{fig:pipeline}
\end{figure}
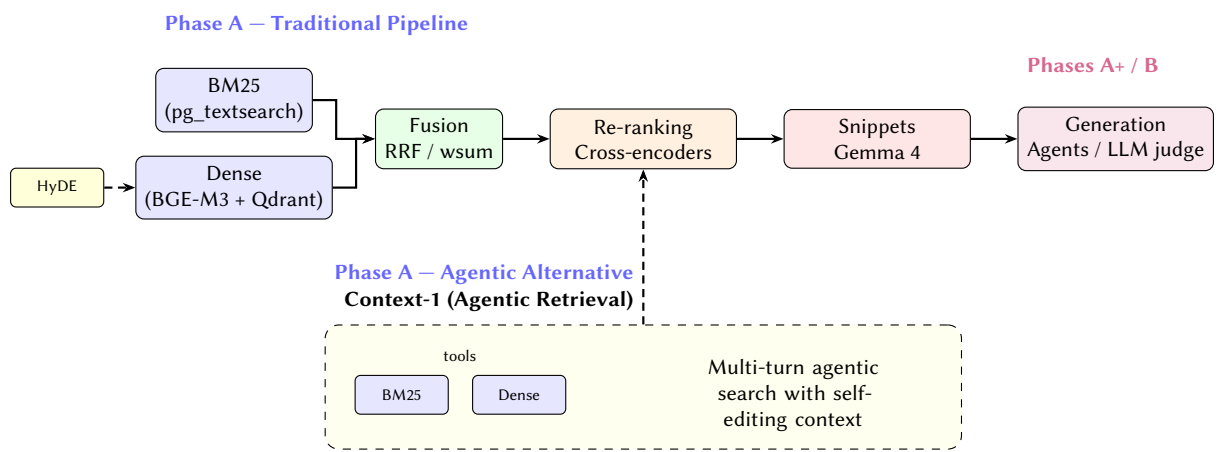

~Figure~\ref{fig:pipeline} provides a visual overview of the complete pipeline. The retrieval process follows a three-stage design: first-stage hybrid retrieval combining BM25 and dense embeddings (the latter with an optional query expansion step using HyDE), followed by neural re-ranking using cross-encoder models. We also investigated an agentic based retrieval system, Context-1 \cite{context1} for LLM-driven retrieval, as an alternative to hybrid retrieval. For snippet generation, a separate fine-tuned model extracts relevant text spans from the top-ranked documents. The fusion of BM25 and dense retrieval results is performed via either Reciprocal Rank Fusion (RRF) or weighted sum, producing a consolidated ranking that feeds into the re-ranking stage. For Phases~A+ and B, the top-ranked documents and their snippets are passed to the agent quorum or LLM-as-judge framework for answer generation.

\subsection{First-Stage Retrieval}
\label{sec:first-stage}

The first stage of our retrieval pipeline performs parallel lexical and dense search over the PubMed corpus, then fuses the results into a single candidate list for subsequent re-ranking. In previous years we relied solely on BM25 as a first stage retrieval model, however this year we experimented with hybrid retrieval combining both dense retrieval with BM25.

\subsubsection{BM25 Retrieval via pg\_textsearch}

In previous editions, we relied on the PISA indexing framework~\cite{mallia2019pisa} accessed through the PyTerrier interface for BM25-based first-stage retrieval. While effective, this approach required loading indexes from disk (slow by design) to memory at each pipeline iteration and maintaining them as separate artifacts from document storage.

For the 14th edition, we migrated BM25 retrieval to PostgreSQL using the pg\_textsearch extension~\cite{pgtextsearch}, an external module that provides full-text search capabilities. This change integrated document storage and full-text search into a single source of truth, eliminating the need for separate index files. The PostgreSQL GiN (Generalized Inverted Index) indexes are maintained automatically alongside document updates, and the entire pipeline now operates on a single database. The BM25 index was configured with parameters $k_1 = 0.4$ and $b = 0.3$, values determined empirically on held-out data. The key improvement is in usability and maintainability: document ingestion, index updates, and retrieval queries all happen within the same system, reducing operational complexity and the risk of synchronization errors between document versions and retrieval indexes.

\subsubsection{Dense Retrieval with Qdrant and BGE-M3}

This year we introduced dense retrieval as a first-stage retrieval method, moving beyond our previous use of dense embeddings solely for Pseudo Relevance Feedback. For embedding generation, we deployed the BGE-M3 model~\cite{chen2024bge} (569M parameters) via the Text Embeddings Interface (TEI) from Hugging Face. BGE-M3 is a multi-lingual, multi-functional embedding model that produces dense vectors suitable for semantic similarity search in the biomedical domain.

TEI is a Rust-based implementation optimized for serving text embeddings. A key architectural advantage of TEI is its API-based design, which decouples embedding generation from the main pipeline process. This allows deploying multiple TEI instances in parallel, scaling horizontally to meet the throughput demands of indexing the full PubMed corpus. TEI provides tokenization and model inference in a single binary supporting both CPU and GPU execution.

The generated embeddings are indexed and stored in Qdrant, a vector database that supports GPU-accelerated approximate nearest neighbor search via HNSW (Hierarchical Navigable Small World) indexing. This replaces our previous approach of precomputing and storing all cross-embedding similarities as flat NumPy arrays---a practice that became increasingly untenable as the document corpus grew. Qdrant provides compact storage, configurable precision-speed tradeoffs, and dynamic collection management with metadata filtering, enabling efficient incremental updates as new documents are indexed.

At query time, the question text (or its HyDE expansion, described below) is encoded via the same TEI endpoint and searched against the Qdrant index. The top 100 dense-retrieved documents are then combined with the top 100 BM25-retrieved documents through fusion.

\subsubsection{Hybrid Fusion}

To combine the BM25 and dense retrieval results into a single candidate list, we evaluated two approaches, weighted sum and Reciprocal Rank Fusion (RRF). The weighted sum method combines scores linearly:

\begin{equation}
  S_{ws}(d) = \alpha \cdot S_{bm25}(d) + (1 - \alpha) \cdot S_{dense}(d)
\end{equation}

where $S_{bm25}(d)$ and $S_{dense}(d)$ are normalized scores from each retrieval method and $\alpha \in [0,1]$ controls the relative contribution of each. We used $\alpha = 0.6$, giving slightly more weight to BM25, which was found to perform well on validation data. While straightforward, this method is sensitive to the scale and distribution of the individual scores, since BM25 and dense similarity scores have different ranges and statistical properties.

Reciprocal Rank Fusion (RRF) provides an alternative that operates on ranks rather than raw scores, making it robust to heterogeneous score distributions. We discuss RRF in the context of re-ranking fusion in Section~\ref{sec:fusion}, as it is primarily used to fuse the outputs of multiple rerankers. For first-stage fusion, both weighted sum and RRF were evaluated; their comparative performance is reported in Section~\ref{sec:results}.

\subsubsection{Query expansion: HyDE}
\label{sec:query-expansion}

Query expansion improves information retrieval by augmenting the initial search with related terms or context to capture a broader intent. Hypothetical Document Embeddings (HyDE) replaces the original query with a hypothetical answer generated by an LLM. Rather than embedding the user's question directly, the LLM is prompted to produce a plausible answer, and this generated text is embedded via TEI and used as the query for dense retrieval. The intuition is that an LLM-generated answer, even if factually incomplete or partially incorrect, shares the linguistic and conceptual structure of genuinely relevant documents more closely than the original question does. When HyDE is enabled, the dense retrieval branch of the first-stage pipeline uses the HyDE embedding instead of the raw question embedding; the BM25 branch continues to use the original query text.

\subsubsection{Agent retrieval: Context-1}

Context-1~\cite{context1} is a 20B-parameter model purpose-trained for multi-turn, agentic search over document collections. Unlike single-pass retrieval pipelines or query expansion methods, Context-1 operates as a retrieval subagent that iteratively decomposes a high-level question into sub-queries, retrieves documents, evaluates their relevance, and refines its search strategy across multiple turns.

The model actively manages its own context window. As it accumulates documents over successive retrieval turns, it selectively discards irrelevant passages through a pruning mechanism, freeing context capacity for further exploration. This self-editing behavior addresses a key limitation of multi-turn retrieval: the tendency of the context window to fill with tangential material. The agent interacts with the underlying search infrastructure through structured tool calls---searching the corpus, reading full documents, and pruning irrelevant chunks---within a fixed token budget that forces explicit relevance judgments.

In our pipeline, Context-1 serves as a retrieval subagent that receives the original BioASQ question and iteratively searches our indexed PubMed collection (via BM25 and dense search tools), internally using the BGE-reranker-v2-m3 cross-encoder to score retrieved candidates. The output of Context-1 is a ranked set of documents, which we then feed through our own ensemble of rerankers for the final ranking. Since Context-1 is a recent project, only the model was available at the time, where we implemented a harness based on the description in the original paper. We implemented the necessary tool interfaces for BM25 and dense search, as well as the pruning mechanism to manage the context window.

\subsection{Re-Ranking}
\label{sec:reranking}


We developed a new, flexible training pipeline for cross-encoder rerankers. The pipeline supports multiple base models, training strategies, and data configurations. A notable improvement was the incorporation of dense retrieval for negative sampling. In previous editions, negative examples for reranker training were selected exclusively from BM25 results, which limited the diversity of negatives. By including dense retrieval results in the negative generation process, the reranker is exposed to a broader range of challenging distractors, improving its ability to distinguish relevant from irrelevant documents across different retrieval paradigms.

We trained a total of 29 reranker models based on diverse architectures including BGE-reranker-v2-m3, PubMedBERT, BioLinkBERT, BioBERT, MedCPT, and the LLaMA Nemotron reranker, among others. Training employed a pairwise loss with varying configurations of epochs, samplers, and hard negative counts. A complete listing of the trained models and their configurations is provided in Appendix~\ref{sec:appendix-models}.

\subsubsection{Sampling strategies}

The training pipeline supports multiple negative sampling strategies. We focus on the two used in our final models.

\textbf{BasicSampler.} This sampler implements straightforward random selection. A positive document is chosen uniformly from any relevance group above the negative threshold (i.e., any document annotated as relevant to the question). A negative document is sampled uniformly from the pool of documents labeled as irrelevant (relevance group~0). When constructing a positive-negative pair, the sampler selects the positive from a randomly chosen relevance tier and the negative from a lower-ranked tier, ensuring that the negative has a strictly weaker relevance signal than the paired positive. This strategy provides a clean training signal but does not differentiate between easy and hard negatives.

\textbf{ShifterSampler.} This sampler implements a curriculum learning strategy that progressively exposes the model to harder negatives as training advances. BM25-retrieved negatives are assumed to be roughly ordered by semantic proximity to the query, with the top-ranked documents (index~0) being the hardest to distinguish from genuine positives. In early epochs, the sampler draws negatives from the full candidate pool $[0, N]$, giving the model a mix of easy and hard examples. As training progresses through later epochs, the sampling window shrinks toward the hard end, eventually sampling only from the top~10 hardest negatives. Given $max\_epoch$ total epochs, at epoch $e$ the window spans $[0, N - \Delta e]$ where $\Delta = N / (max\_epoch + 1)$. This forces the reranker to learn increasingly fine-grained distinctions between relevant documents and their most similar distractors, mirroring the challenge it faces in production where the top-ranked BM25 candidates are precisely the documents most likely to be confused with relevant ones.

\subsubsection{Reciprocal Rank Fusion (RRF)}
\label{sec:fusion}

To combine the outputs of multiple rerankers into a single ranked list, we employ Reciprocal Rank Fusion (RRF). RRF combines rankings based solely on reciprocal positions rather than raw scores, making it robust to score scale differences between heterogeneous models. The RRF score for a document $d$ is computed as:

\begin{equation}
  RRF(d) = \sum_{r \in R} \frac{1}{k + rank_r(d)}
\end{equation}

where $R$ is the set of reranker runs and $k$ is a constant (typically 60) that mitigates the impact of high-ranked documents dominating the final score. Because RRF operates on ranks rather than scores, it requires no normalization and is insensitive to the magnitude of individual reranker scores. This is particularly important when combining rerankers of different architectures (e.g., BERT-based and decoder-based models) whose score distributions differ substantially. In our experiments, RRF consistently outperformed weighted sum for reranker fusion, as the rank-based approach naturally handles heterogeneous score scales without requiring per-batch calibration.

\subsection{Snippet Generation via Fine-Tuned Gemma 4}

For the first time, we participated in the snippet generation subtask. Given a question and a set of retrieved documents, the system must identify the most relevant text spans within those documents. We approached this as a text generation problem: a model receives the question along with a document's title and abstract, and generates the most relevant snippets from that document.

We fine-tuned a Gemma~4~31B model using QLoRA (Quantized Low-Rank Adaptation) for this task. The training data was derived from the snippet annotations provided in the BioASQ training set, where for each question-document pair, the gold-standard snippets were extracted from the document text. The model was trained in a chat format, receiving the question and document content as input and producing the corresponding snippets as output. The LoRA adapters were applied to the attention projection layers, enabling efficient fine-tuning with substantially reduced memory requirements compared to full fine-tuning.


\subsection{Answer Generation}

Once relevant documents and snippets have been retrieved, the system must produce a final answer to the biomedical question. Rather than relying on a single generation call, this edition explores two complementary strategies for combining and refining candidate outputs from multiple models: an LLM-as-a-judge framework that evaluates and selects among independently generated answers, and an agent quorum mechanism that has multiple models debate and converge on a shared answer. Both approaches aim to improve robustness over simple majority voting by incorporating richer signals about correctness, evidentiary support, and completeness. The following subsections describe each in detail.

\subsubsection{Majority Voting}

For Batch 1, we employed previous editions' majority voting as the primary method for aggregating outputs from multiple generation models. Each model produces an answer independently given the same question and context, and the final answer is selected by a fixed, rule-based aggregation strategy that varies by question type. For yes/no questions, a simple majority vote is taken: whichever answer (``yes'' or ``no'') appears most frequently across model outputs wins, with ties defaulting to ``yes.'' For factoid questions, which ask for a specific named entity, models produce ranked lists of up to five candidates. These lists are merged via Reciprocal Rank Fusion (RRF): each candidate receives a score equal to the sum of $1 / (rank + k)$ across all input lists, with $k = 60$, and the top five candidates by fused score are returned. For list questions, entities that appear in at least a threshold fraction of the input sets (defaulting to a strict majority, i.e., 50\% of models) are retained in the final output. This approach is straightforward and computationally inexpensive, but it treats all models as equally reliable and provides no mechanism for evaluating whether a candidate answer is actually supported by the retrieved evidence.

\subsubsection{LLM as a Judge}
\label{sec:llm-judge}

For this edition, we replaced majority voting with an LLM-as-a-judge framework. Rather than applying a fixed aggregation rule, a separate LLM instance evaluates the outputs of different generation models and either selects the best candidate answer or synthesizes a superior answer by combining elements from multiple outputs. The judge model receives all candidate answers along with the original question, the question type, the retrieved context, and (when available) reference ideal answers. It is prompted to score each candidate answer along multiple dimensions (correctness, faithfulness to the provided context, and completeness) on a 0--1 scale, with an accompanying rationale.

The judge's system prompt establishes it as a biomedical question-answering evaluator that outputs a structured JSON object containing numerical scores and a concise rationale. As stated before, the scoring dimensions are: \emph{correctness} (factual accuracy of the answer), \emph{faithfulness} (whether claims are grounded in the retrieved context), \emph{completeness} (whether all aspects of the question are addressed), and \emph{overall} (a holistic quality judgment). This multi-dimensional evaluation provides a richer signal than binary voting and allows the pipeline to prefer answers that are well-supported by evidence, even if they are not the most similar to the majority opinion.

The motivation for adopting an LLM judge over majority voting is twofold. First, an LLM can critically evaluate the evidence grounding of each answer, penalizing hallucinations that a voting scheme would not detect. Second, the judge can identify complementary information across different candidate answers and either select the most comprehensive one or, in some configurations, synthesize a hybrid answer. This flexibility is particularly valuable for ideal answer generation, where different models may emphasize different aspects of the evidence.

In practice, the LLM judge proved especially effective for exact answer types (yes/no, factoid, list), where correctness is more objectively verifiable against context.

\subsubsection{Agent Quorum Mechanism}

The most significant methodological innovation for this edition is the agent quorum, a multi-agent debate framework designed to produce more robust and well-reasoned answers. The mechanism operates as follows.

\textbf{Agent configuration and thinking focuses.} Multiple agents are instantiated, each with a potentially different LLM and a distinct cognitive focus that shapes how they approach the evidence. The available focuses, drawn from a predefined pool, include: \emph{analytical} (systematically decomposing evidence and identifying logical chains), \emph{evidence-based} (grounding every claim strictly in the provided documents and flagging contradictions between sources), \emph{skeptical} (challenging the debate's direction, seeking gaps in evidence, alternative interpretations, and methodological limitations), \emph{integrative} (synthesizing information across all sources and reconciling apparent contradictions), \emph{pragmatic} (focusing on the most direct, actionable answer and prioritizing precision), and \emph{theoretical} (considering underlying biological mechanisms and established scientific principles). Each agent receives a system prompt that establishes it as a rigorous biomedical scientist participating in a structured expert debate, responding exclusively with valid JSON.

\textbf{Adaptive document retention.} Documents and their corresponding snippets are provided to agents through an adaptive retention protocol. Initially, each agent receives three randomly selected documents from the retrieval set. After analyzing these documents, each agent decides which documents to retain. If an agent retains fewer than three documents, its next round includes those retained documents plus additional randomly selected documents to bring the total back to three. If an agent retains all three, the next round includes four documents (the three retained plus one new random document), and so on. This mechanism allows agents to maintain focus on the documents they deem most relevant while preserving some diversity in the evidence pool, preventing premature convergence on suboptimal information.

\textbf{Debate and scoring.} In each round, every agent produces both an opinion (its current best answer to the question with detailed reasoning) and an agreement level drawn from the set $\{\text{strongly\_disagree}, \text{disagree}, \text{agree}, \text{strongly\_agree}\}$. Agents are ordered randomly in each round to prevent anchoring effects. During the debate, agents are shown the opinions and agreement levels of all other agents from the current round and are instructed to engage critically: if another agent cites evidence they have not seen, they weigh those claims against their own documents; if they disagree, they explain what specific evidence or reasoning undermines the opposing position; if they agree, they strengthen the argument with additional evidence from their documents. Agents are explicitly warned against premature agreement and are instructed to output \texttt{strongly\_agree} only when genuinely confident that the answer is correct, complete, and well-supported by evidence.

\textbf{Convergence criteria.} The quorum continues for multiple rounds until one of three stopping conditions is met: (1)~all agents reach the \texttt{strongly\_agree} level, indicating full consensus; (2)~a predefined maximum number of rounds is reached; or (3)~the agreement levels remain unchanged for a specified number of consecutive rounds (default: 3), indicating a stable disagreement that further debate is unlikely to resolve.

\textbf{Final summarization.} Once the quorum concludes, a designated synthesizer model consolidates all debate turns into a single final answer. This model receives the original question, the full set of context documents, and a structured summary of the debate. It is instructed to produce a definitive answer based solely on the evidence in the context documents, informed by the reasoning developed during the debate, without referencing the debate process itself.

\textbf{Agent configurations across batches.} The models and thinking focuses used in the agent quorum evolved across batches as we experimented with different configurations. Table~\ref{tab:quorum-batches} summarizes the agent configurations used in each test batch.

\begin{table}[ht]
  \centering
  \caption{Agent quorum configurations across test batches. Agents are listed with their model identifier and the synthesizer used for final answer generation. The \textbf{Config} code encodes the number of agents, weight openness (\texttt{open} = all weights public, \texttt{mixed} = some API-only), and the parameter range (in billions) across open-weight agents; \texttt{api} denotes configurations dominated by hosted closed models.}
  \label{tab:quorum-batches}
  \begin{tabularx}{\textwidth}{llX}
    \toprule
    \textbf{Batch} & \textbf{Version / Config} & \textbf{Agents (Model)} \\
    \midrule
    \multirow{8}{*}{B2}& V1 (\texttt{3-open-27-120b})      & Qwen3-Next-80B, Nemotron-3-Super-120B, MedGemma-27B \\
    &                                      & Synthesizer: Nemotron-3-Super-120B \\
    & V2 (\texttt{5-mixed-api})                & Nemotron-3-Super-120B, Grok-4.1-Fast, Gemini-3-Flash, Gemini-2.5-Flash, GPT-5-Mini \\
    &                                      & Synthesizer: Claude-Sonnet-4.6 \\
    & V3 (\texttt{4-open-35-309b})         & Qwen3-Next-80B, Mimo-v2-Flash (309B), Qwen3.5-35B, Nemotron-3-Super-120B \\
    &                                      & Synthesizer: GPT-5.4-Nano \\
    & V4 (\texttt{3-gemma-26-31b})          & Gemma-4-31B, Gemma-4-26B, MedGemma-27B \\
    &                                      & Synthesizer: Gemma-4-31B \\
    \midrule
    \multirow{8}{*}{B3} & V0 (\texttt{3-open-2-4b})         & Gemma-4-E4B, Gemma-4-E2B, Nemotron-3-Nano-4B \\
    &                                      & Synthesizer: Gemma-4-E4B \\
    & V1 (\texttt{4-open-26-35b})          & Gemma-4-26B, Gemma-4-31B, Nemotron-3-Nano-30B, Qwen3.5-35B \\
    &                                      & Synthesizer: Gemma-4-26B \\
    & V2 (\texttt{5-mixed-api})            & Nemotron-3-Super-120B, Grok-4.1-Fast, Gemini-3-Flash, Gemini-2.5-Flash, GPT-5-Mini \\
    &                                      & Synthesizer: Gemini-2.5-Flash \\
    & V3 (\texttt{4-open-24-675b})         & Gemma-4-31B, Qwen3.5-35B, Mistral-Small-4 (119B), Mistral-Large-3 (675B) \\
    &                                      & Synthesizer: Mistral-Small-4 (119B) \\
    & V4 (\texttt{3-mistral-129-695b})     & Mistral-Medium-3.5 (128B), Mistral-Small-4 (119B), Mistral-Large (695B) \\
    &                                      & Synthesizer: Mistral-Medium-3.5 (128B) \\

    \midrule
    \multirow{10}{*}{B4} & V0 (\texttt{3-open-2-4b})         & Gemma-4-E4B, Gemma-4-E2B, Nemotron-3-Nano-4B \\
    &                                      & Synthesizer: Gemma-4-E4B \\
    & V1 (\texttt{2-gemma-4-26-31b})       & Gemma-4-26B, Gemma-4-31B \\
    &                                      & Synthesizer: Gemma-4-26B \\
    & V2 (\texttt{4-mixed-api})            & Nemotron-3-Super-120B, Gemini-3-Flash, Gemini-2.5-Flash, GPT-5-Mini \\
    &                                      & Synthesizer: Gemini-2.5-Flash \\
    & V3 (\texttt{3-qwen-27-35b})    & Qwen3.6-35B, Qwen3.6-27B, Qwen3.6-Flash \\
    &                                      & Synthesizer: Qwen3.6-35B \\
    & V4 (\texttt{3-mistral-129-695b})     & Mistral-Medium-3.5 (128B), Mistral-Small-4 (119B), Mistral-Large (695B) \\
    &                                      & Synthesizer: Mistral-Medium-3.5 (128B) \\
    & V5 (\texttt{2-mistral-120B})         & Mistral-Medium-3.5 (128B), Mistral-Small-4(119B) \\
    &                                      & Synthesizer: Mistral-Medium-3.5 (128B)\\
    \bottomrule
  \end{tabularx}
\end{table}

\section{Results}
\label{sec:results}

This section presents both our internal validation results and the official evaluation results for each phase across the four test batches of the 14th BioASQ challenge.

\subsection{Validation}

These experiments were conducted on the 13B Batch~1--4 golden datasets and were designed to measure the marginal contribution of each retrieval component to end-to-end retrieval performance.

\subsubsection{First Stage Retrieval}
\label{sec:internal-retrieval}

We conducted extensive internal testing of the various retrieval strategies introduced this year, including Context-1-based retrieval and different fusion approaches (RRF and weighted sum) with and without neural reranking. Table~\ref{tab:internal-retrieval} presents the results using standard retrieval metrics: M@100 (Match@100, average number of relevant documents in the top 100), H@100 (Hit@100, fraction of questions with $\ge$1 relevant document), AR@100 (AllRel@100, fraction of all relevant documents recovered), MRR (Mean Reciprocal Rank), and R@100 (Recall@100). MRR is the primary metric for ranking systems in Phase~A, while the others provide complementary insights into retrieval effectiveness. R@100 and AR@100 are particularly important for Phase~A+ and Phase~B, where the quality of retrieved documents directly impacts answer generation performance.

\begin{table}[ht]
  \centering
  \caption{Internal retrieval testing results comparing BM25 baseline, fusion strategies of BM25 and dense retrieval (hybrid RRF and weighted sum), reranked variants, and Context-1. Bold values indicate the best result in each column.}
  \label{tab:internal-retrieval}
  \footnotesize
  \footnotesize
  \begin{tabular}{l|cccccc}
    \toprule
    \textbf{Method} & \textbf{Docs} & \textbf{M@100} & \textbf{H@100} & \textbf{AR@100} & \textbf{MRR} & \textbf{R@100} \\
    \midrule
    BM25 (baseline) & 99.3 & 11.01 & 0.953 & 0.082 & 0.549 & 0.373 \\
    \midrule
    Hybrid RRF & 100.0 & 12.28 & 0.965 & 0.047 & 0.696 & 0.406 \\
    Hybrid RRF + BGE-reranker-base & 100.0 & 12.28 & 0.965 & 0.047 & 0.498 & 0.406 \\
    Hybrid RRF + BGE-reranker-v2-m3 & 100.0 & 12.28 & 0.965 & 0.047 & 0.649 & 0.406 \\
    \midrule
    Hybrid weighted sum & 100.0 & 11.47 & 0.965 & 0.082 & 0.622 & 0.384 \\
    Hybrid wsum + BGE-reranker-base & 100.0 & 11.47 & 0.965 & 0.082 & 0.597 & 0.384 \\
    Hybrid wsum + BGE-reranker-v2-m3 & 100.0 & 11.47 & 0.965 & 0.082 & \textbf{0.738} & 0.384 \\
    \midrule
    Context-1 & 100.0 & \textbf{14.16} & 0.965 & 0.047 & 0.508 & \textbf{0.453} \\
    \bottomrule
  \end{tabular}
\end{table}

Several patterns emerge from these results. The BM25 baseline achieves the lowest M@100 (11.01), MRR (0.549), and R@100 (0.373), confirming that lexical retrieval alone is insufficient for this task. Hybrid retrieval, whether fused via RRF or weighted sum, substantially improves over BM25 across all metrics: MRR increases from 0.549 to 0.696 (RRF) and 0.622 (weighted sum), while R@100 rises from 0.373 to 0.406 and 0.384, respectively. RRF consistently outperforms weighted sum on MRR and Recall, consistent with its robustness to heterogeneous score distributions.

The addition of a neural reranker produces mixed but instructive results. When applied to the RRF-fused results, the BGE-reranker-v2-m3 yields an MRR of 0.649, slightly below the un-reranked RRF baseline (0.696), while the base variant drops to 0.498. However, when the reranker is applied to the weighted-sum results, the v2-m3 model achieves the best MRR overall (0.738), suggesting an interaction between the fusion method and the reranker: weighted sum preserves score magnitudes that the reranker can exploit, whereas RRF discards score information in favor of ranks. The base reranker variant consistently underperforms its v2-m3 counterpart, confirming the expected performance hierarchy between these models observed in our validation.

Context-1 achieves the highest M@100 (14.16) and R@100 (0.453), demonstrating its ability to surface a broader set of relevant documents than the single-pass hybrid approaches. However, its MRR (0.508) is lower than the best fusion configurations, indicating that while Context-1 is effective at exploration---finding more relevant documents overall---it is less precise at ranking the most relevant documents first. This is consistent with Context-1's design as a retrieval subagent optimized for recall, intended to feed a downstream reranker or reasoning model rather than to produce a final ranking directly.

As such, some choices in our official submissions were guided by these insights: we prioritized the RRF and weighted-sum hybrid approaches for its superior MRR when combined with the BGE-reranker-v2-m3, while also including Context-1-based runs to leverage its strong recall capabilities, albeit with the understanding that it may require more aggressive reranking to achieve competitive precision.

\subsubsection{Reranker Models}
\label{sec:internal-validation}

We conducted internal validation of our reranker models using the 13B golden datasets (batches~1--4) as held-out evaluation sets. Performance was measured using the BioASQ-specific MAP metric (MAP@10, based on PubMed article-level relevance judgments). Table~\ref{tab:reranker-val} summarizes the validation results for the top-performing models.

\begin{table}[ht]
  \centering
  \caption{Internal validation results for selected reranker models on the 13B golden datasets, sorted by performance.}
  \label{tab:reranker-val}
  \footnotesize
  \begin{tabular}{lcc}
    \toprule
    \textbf{Model} & \textbf{Configuration} & \textbf{MAP-bioasq@10} \\
    \midrule
    LLaMA Nemotron rerank 1B & E2-S4, multi\_neg\_pairwise, InfoNCE & 0.9995 \\
    LLaMA Nemotron rerank 1B & E5, pairwise & 0.9970 \\
    BGE-reranker-v2-m3 & E5, pairwise & 0.6824 \\
    BGE-reranker-base & E5, pairwise & 0.6686 \\
    nboost/pt-biobert-base-msmarco & E5, pairwise & 0.6608 \\
    ncbi/MedCPT-Cross-Encoder & E5, pairwise & 0.6404 \\
    BioLinkBERT-base & E5, pairwise & 0.6403 \\
    cross-encoder/ms-marco-MiniLM-L-6-v2 & E5, pairwise & 0.6373 \\
    PubMedBERT-base & E5, pairwise & 0.6291 \\
    \bottomrule
  \end{tabular}
\end{table}

The LLaMA Nemotron reranker models achieved outstanding validation performance, substantially outperforming all BERT-based architectures. There are, however, concerns of data leakage due to the unusually high score. Among the smaller models, BGE-reranker-v2-m3 demonstrated the strongest performance, outperforming the model used in previous years (BioLinkBERT-Base), followed by the biomedical domain-specific models. The gap between general-domain and biomedical-specific rerankers was narrower than in previous years, suggesting that larger pretrained models can effectively transfer to biomedical text without extensive domain-specific pretraining.

\subsubsection{Sampling strategies}

We also evaluated the impact of the negative sampling strategy on reranker quality. Table~\ref{tab:reranker-sampling-val} compares the BasicSampler (uniform random negatives) against the ShifterSampler (curriculum learning with progressively harder negatives), described in Section~\ref{sec:reranking}. Models were trained with identical architectures, epochs, and data (differing only in the sampler) to isolate the contribution of the sampling strategy.

\begin{table}[ht]
  \centering
  \caption{Internal validation results for sampling strategies to reranker training, sorted by performance.}
  \label{tab:reranker-sampling-val}
  \footnotesize
  \begin{tabular}{lcc}
    \toprule
    \textbf{Model} & \textbf{Configuration} & \textbf{MAP-bioasq@10} \\
    \midrule
    \multirow{2}{*}{BGE-reranker-v2-m3}
    & basic & 0.6489 \\
    & shifter & 0.6824 \\
    \midrule
    \multirow{2}{*}{BGE-reranker-base}
    & basic & 0.6312 \\
    & shifter & 0.6686 \\
    \midrule
    \multirow{2}{*}{cross-encoder/ms-marco-MiniLM-L-6-v2}
    & basic & 0.5928 \\
    & shifter & 0.6373 \\
    \midrule
    \multirow{2}{*}{PubMedBERT-base}
    & basic & 0.6239 \\
    & shifter & 0.6291 \\
    \bottomrule
  \end{tabular}
\end{table}

The ShifterSampler consistently outperformed the BasicSampler across all four model architectures, with absolute MAP improvements ranging from 0.0052 (PubMedBERT-base) to 0.0445 (MiniLM-L-6-v2). The largest relative gain was observed on the smallest model (MiniLM-L-6-v2, 67M parameters), however, even on larger architectures such as BGE-reranker-v2-m3, the shifter contributed a 0.0335 MAP improvement over the basic sampler.

\subsubsection{Answer Generation}

For this phase, internal validation is considered unreliable, as there are no direct, targeted metrics that correlate well with human judgments of answer quality. Instead, we performed qualitative analysis of generated answers using the 13B golden datasets, focusing on the consistency of the multi-agent quorum mechanism and the impact of adaptive document retention. We observed that while the agent quorum often converged on a consensus answer, there were cases where agents with diverse prompts produced conflicting outputs, leading to less coherent final answers. Adaptive document retention based on snippet-level evidence showed promise in improving answer relevance, but its effectiveness varied depending on the question type and the quality of retrieved snippets. These insights informed our final system configurations for the official submissions.

\subsection{Official Results}
Our systems are labeled \texttt{bioinfo-0} through \texttt{bioinfo-4}, representing different configurations submitted across batches. For each absolute-value table, we also provide the corresponding ranking position within the full competition field (lower rank = better). Systems sharing the same score receive the same rank, and subsequent ranks are skipped accordingly.

\subsubsection{Document Retrieval (Phase A)}
\label{sec:results-phase-a}
The official Phase~A document retrieval results across the four test batches are presented in Table~\ref{tab:phasea-all}. In Batch 1, we utilized BM25 and DPRF exclusively for first-stage retrieval. We primarily experimented with new reranker models by first filtering them (bioinfo-0 and bioinfo-1), evaluating the complete set (bioinfo-2), and employing the previous year's rerankers (bioinfo-3). The key takeaway from these runs was that the filtered models performed significantly worse, indicating that the Nemotron models may not be performing as well as initially anticipated. Additionally, we noted that the 2025 rerankers outperformed the 2026 variants. Upon further investigation, we discovered that this batch suffered from incomplete indexing, with approximately one million missing documents. We believe this omission severely impacted model efficacy and degraded first-stage retrieval performance for this batch, an issue subsequently corrected in Batch 2.

For Batch 2, we began experimenting with hybrid retrieval. Examining runs bioinfo-0, bioinfo-2, and bioinfo-4—where the top-performing rerankers were paired with varying retrievers—we observe that hybrid retrieval using a weighted sum substantially outperforms alternative approaches. Conversely, hybrid Reciprocal Rank Fusion (RRF) underperformed relative to the BM25 baseline. Several factors could account for these outcomes; most notably, these trends may shift following the release of the final official judgments. Alternatively, this pattern may imply that BM25 features carry higher relative importance than dense retrieval in this specific context. However, looking at bioinfo-1 (which incorporated all rerankers), we observed nearly identical performance to our best system, whilst using a BM25-only configuration, whereas bioinfo-3 experienced a performance drop, suggesting a potential processing anomalies within this specific batch.

In Batch 3, we investigated query expansion techniques via bioinfo-1 and bioinfo-2; however, these configurations once again fell short of the established BM25 baseline. We also evaluated the Context-1 model for first-stage retrieval, which yielded negative results. We again observed that the weighted-sum hybrid approach outperformed RRF hybrid retrieval. Unfortunately, the absence of strict control tests prevents us from verifying whether query expansion provided any intrinsic benefits or merely introduced unnecessary computational overhead.

For Batch 4, we maintained a similar submission strategy but replaced the weighted-sum hybrid run with a control RRF hybrid run to benchmark against a HyDE submission. The results indicate that HyDE significantly boosts retrieval effectiveness, marginally outperforming the BM25 baseline in this instance.

Overall, the results this year appear worse than in previous iterations, with our most successful submissions closely mirroring our prior-year architectures. This suggests that either the newly introduced hybrid configurations are less effective, our evaluation baseline is inherently biased toward BM25, or performance metrics will normalize once final ground-truth evaluations are released.

\begin{table}[ht]
  \centering
  \caption{Phase~A document retrieval results across all four Test Batches. Best is the top competitor MAP (rank 1); Median is the median competitor MAP for that batch.}
  \label{tab:phasea-all}
  \footnotesize
  \begin{tabular}{cl|lll|cc}
    \toprule
    \textbf{Batch} & \textbf{System} & \textbf{Retriever} & \textbf{Reranker Ensemble} & \textbf{N} & \textbf{MAP} & \textbf{Rank} \\
    \midrule
    \multirow{7}{*}{1}
    & bioinfo-0 & BM25 & Filtered (MAP$>$0.64) & 5--7 & 0.1476 & 34 \\
    & bioinfo-1 & BM25 + DPRF & Filtered (MAP$>$0.64) & 5--7 & 0.1502 & 30 \\
    & bioinfo-2 & BM25 + DPRF & All 2026 rerankers & $\sim$24 & 0.2220 & 14 \\
    & bioinfo-3 & BM25 + DPRF & All 2025 rerankers & $\sim$46 & \textbf{0.2479} & \textbf{5} \\
    & bioinfo-4 & BM25 + DPRF & All 2025 + 2026 rerankers & $\sim$70 & 0.2398 & 11 \\
    \cmidrule(l){2-7}
    & \textit{Best} & \multicolumn{3}{l|}{} & 0.2835 & 1 \\
    & \textit{Median} & \multicolumn{3}{l|}{} & 0.1691 & 27 \\
    \midrule
    \multirow{7}{*}{2}
    & bioinfo-0 & BM25 & Best rerankers & $\sim$5 & 0.0813 & 59 \\
    & bioinfo-1 & BM25 & All 2026 rerankers & $\sim$29 & 0.1914 & 15 \\
    & bioinfo-2 & Hybrid RRF & Best rerankers & $\sim$5 & 0.0721 & 60 \\
    & bioinfo-3 & Hybrid RRF & All 2026 rerankers & $\sim$29 & 0.0718 & 61 \\
    & bioinfo-4 & Hybrid wsum & Best rerankers & $\sim$5 & \textbf{0.1922} & \textbf{14} \\
    \cmidrule(l){2-7}
    & \textit{Best} & \multicolumn{3}{l|}{} & 0.2395 & 1 \\
    & \textit{Median} & \multicolumn{3}{l|}{} & 0.1406 & 33--34 \\
    \midrule
    \multirow{7}{*}{3}
    & bioinfo-0 & BM25 & Best rerankers (baseline) & $\sim$5 & \textbf{0.1887} & \textbf{5} \\
    & bioinfo-1 & Hybrid wsum + HyDE & Best rerankers & $\sim$5 & 0.1600 & 17 \\
    & bioinfo-2 & Hybrid RRF + HyDE & Best rerankers & $\sim$5 & 0.1419 & 21 \\
    & bioinfo-3 & Context-1 & 2025 rerankers & $\sim$46 & 0.1088 & 47 \\
    & bioinfo-4 & Context-1 & 2026 rerankers & $\sim$24 & 0.0993 & 49 \\
    \cmidrule(l){2-7}
    & \textit{Best} & \multicolumn{3}{l|}{} & 0.2114 & 1 \\
    & \textit{Median} & \multicolumn{3}{l|}{} & 0.1253 & 31--32 \\
    \midrule
    \multirow{7}{*}{4}
    & bioinfo-0 & BM25 & Best rerankers (baseline) & $\sim$5 & 0.1951 & 13 \\
    & bioinfo-1 & Hybrid RRF & Best rerankers & $\sim$5 & 0.1163 & 53 \\
    & bioinfo-2 & Hybrid RRF + HyDE & Best rerankers & $\sim$5 & \textbf{0.1959} & \textbf{11} \\
    & bioinfo-3 & Context-1 & Best rerankers & $\sim$5 & 0.1538 & 40 \\
    & bioinfo-4 & Context-1 & All 2026 rerankers & $\sim$24 & 0.1538 & 41 \\
    \cmidrule(l){2-7}
    & \textit{Best} & \multicolumn{3}{l|}{} & 0.2310 & 1 \\
    & \textit{Median} & \multicolumn{3}{l|}{} & 0.1598 & 38--39 \\
    \bottomrule
  \end{tabular}
\end{table}

\subsubsection{Snippet Retrieval (Phase A)}
\label{sec:results-snippets}

For the snippet retrieval subtask, we participated in Batch~3 and Batch~4. Table~\ref{tab:snippets} presents the simplified results showing MAP, rank, and the configuration used for each submission. All snippet submissions used the corresponding Phase~A document retrieval pipeline as their document source, with the same snippet extraction performed by the generation models described in Section~\ref{sec:methodology}.

\begin{table}[ht]
  \centering
  \caption{Snippet retrieval results across test batches. All submissions used the Phase~A retrieval pipeline for document sourcing; snippet extraction was performed by the same generation models used in Phase~A+/B. System ranks are calculated out of all participating systems (50 in B3, 66 in B4).}
  \label{tab:snippets}
  \footnotesize
  \begin{tabular}{lcccc}
    \toprule
    \multirow{2}{*}{\textbf{System}} & \multicolumn{2}{c}{\textbf{Batch B3}} & \multicolumn{2}{c}{\textbf{Batch B4}} \\
    \cmidrule(lr){2-3} \cmidrule(lr){4-5}
    & \textbf{MAP} & \textbf{Rank} & \textbf{MAP} & \textbf{Rank} \\
    \midrule
    bioinfo-0 & \textbf{0.0546} & \textbf{25} & \textbf{0.0591} & \textbf{28} \\
    bioinfo-1 & 0.0352 & 32 & 0.0296 & 51 \\
    bioinfo-2 & 0.0375 & 30 & 0.0495 & 35 \\
    bioinfo-3 & 0.0267 & 38 & 0.0431 & 42 \\
    bioinfo-4 & 0.0130 & 47 & 0.0490 & 36 \\
    \midrule
    \textbf{Best}   & 0.1440 & 1 & 0.1750 & 1 \\
    \textbf{Median} & 0.0543 & 25-26 & 0.0512 & 33-34 \\
    \bottomrule
  \end{tabular}
\end{table}

Snippet retrieval performance was modest: the best MAP ranks were 25th (bioinfo-0, Batch 3) and 28th (bioinfo-0, Batch 4). Interestingly, the BM25 + best rerankers baseline achieved our best snippet rank in both batches. In Batch 3 this is consistent with bioinfo-0 also being our best-ranked document retrieval system that batch; in Batch 4, however, bioinfo-0 ranked 13th in document retrieval, behind bioinfo-2 (rank 11), so the snippet advantage there is not simply inherited from document-retrieval rank. Context-1, despite its strong recall for documents, produced lower snippet MAP...

\subsubsection{Exact Answers (Phases A+ and B)}
\label{sec:results-phase-aplus}

Phase~A+ requires generating exact answers (yes/no, factoid, and list-type questions). For each batch, we present both the absolute scores and the competition rankings side by side. Each system description indicates the models used for exact answer generation, with the source used for documents in phase A+ being always bioinfo-0 from phase A.

For exact answers we report the official metrics: F1 for yes/no questions,
MRR for factoid questions, and F1 for list questions, in Phase~A+
(Table~\ref{tab:phaseap-exact-all}) and Phase~B (Table~\ref{tab:phaseb-exact-all}).

In Phase~A+, our systems were most competitive on the factoid and list metrics. In Batch~1 our submissions clustered around the median, with \texttt{bioinfo-1} and \texttt{bioinfo-2} tied for our best factoid rank (0.3261, rank~13), and \texttt{bioinfo-1} and \texttt{bioinfo-3} tied for our best list rank (0.1735, rank~20). Batch~2 showed a clear benefit from the LLM-as-judge stage: the judge-based runs (best candidate answer at \texttt{bioinfo-3} and top-3 best candidates at \texttt{bioinfo-4}, as described in Section~\ref{sec:llm-judge}) lifted list F1 to 0.3145 (rank~15) and 0.3227 (rank~13), well above the plain agent quorums (0.18--0.25), while also leading on yes/no and factoid. Our strongest batch was Batch~3, where two quorums reached a perfect yes/no F1 of 1.0 (rank~1): the open-source \texttt{bioinfo-1} and the mixed-API \texttt{bioinfo-2}. The open-source \texttt{bioinfo-3} quorum placed 3rd on factoid (0.4706) and 6th on list (0.3138), exceeding the mixed-API \texttt{bioinfo-2} on both metrics and indicating that an all-open-source quorum was competitive with proprietary configurations; its yes/no score (0.8952, rank~18) was above median but below our top systems that batch. In Batch~4, the Mistral quorum \texttt{bioinfo-4} placed 2nd on factoid (0.3636); list F1 stayed above median for two of the three smaller open quorums (\texttt{bioinfo-1}, \texttt{bioinfo-3}), though the smallest configuration, \texttt{bioinfo-0} (v0~[S]), fell below median on this metric (0.2316 vs.\ 0.3065).

Phase~B follows a similar pattern, with differentiation coming from factoid and list
while yes/no scores sit at or slightly below the median. In Batch~1, both submissions
used the LLM-as-judge pipeline, and widening the judge from top-$k$\,=\,3 to
top-$k$\,=\,5 improved factoid MRR from 0.4130 to 0.4565 (rank~8). Batch~2 again
favoured the judge/ensemble systems: LLM-as-judge with ensemble summarization
(\texttt{bioinfo-3}) gave our best list F1 (0.4366, rank~15), while the Sonnet-4.6
ideal-answer ensemble (\texttt{bioinfo-4}) led on factoid (0.3500, rank~10). In
Batch~3 the Mistral quorum (\texttt{bioinfo-4}) reached 0.5000 factoid MRR (rank~8), greatly outperforming the other systems.
Batch~4 was our strongest Phase~B result: the Qwen quorum (\texttt{bioinfo-3}) placed
5th on factoid (0.5000) and 9th on list (0.5671), and two systems (\texttt{bioinfo-0},
\texttt{bioinfo-4}) reached 3rd on yes/no (0.9352).

Across both phases, yes/no F1 was a poor discriminator---scores were tightly bunched
near the median for almost all participants---whereas factoid and list F1 separated
systems more clearly and accounted for nearly all of our top-10 finishes. Two design
choices recur in our best runs: the LLM-as-judge / ensemble-selection stage, which
gave the largest gains on list F1, and open-source quorums that matched or exceeded
mixed proprietary-API configurations.

\begin{table}[ht]
  \centering
  \caption{Phase~A+ exact answer results across all four Test Batches. Scores are shown with their overall leaderboard ranking in parentheses. The top-performing system among our submissions in each batch is highlighted in bold. \textbf{Best} represents the top competitor score (Rank 1); \textbf{Median} is the median score (and median rank) across all participants for that batch.}
  \label{tab:phaseap-exact-all}
  \footnotesize
  \begin{tabular}{cl|p{5.8cm}|ccc}
    \toprule
    \textbf{Batch} & \textbf{System} & \textbf{Generation Configuration} & \textbf{Y/N F1} & \textbf{Fact. MRR} & \textbf{List F1} \\
    \midrule
    \multirow{7}{*}{1}
    & bioinfo-0 & MedGemma-27B + Qwen3.5 35B & 0.8211 (37) & 0.2696 (23) & 0.1509 (28) \\
    & bioinfo-1 & Qwen3.5 235B + Qwen Max Thinking & \textbf{0.8786 (16)} & \textbf{0.3261 (13)} & \textbf{0.1735 (20)} \\
    & bioinfo-2 & Gemini 2.0 Flash + 2.5 Flash + Qwen3.5 122B & \textbf{0.8786 (16)} & \textbf{0.3261 (13)} & 0.1731 (22) \\
    & bioinfo-3 & Gemini 2.0 Flash + 2.5 Flash & \textbf{0.8786 (16)} & 0.2609 (24) & \textbf{0.1735 (20)} \\
    & bioinfo-4 & All models ensemble & \textbf{0.8786 (16)} & 0.2609 (24) & 0.1731 (22) \\
    \cmidrule(l){2-6}
    & \textit{Best} & \multicolumn{1}{l|}{} & 1.0000 (1) & 0.4783 (1) & 0.2345 (1) \\
    & \textit{Median}& \multicolumn{1}{l|}{} & 0.8786 (27) & 0.2761 (27) & 0.1731 (27) \\
    \midrule
    \multirow{7}{*}{2}
    & bioinfo-0 & Agents v1 (3-open-27-120b) & 0.8444 (28) & 0.2000 (39) & 0.1778 (48) \\
    & bioinfo-1 & Agents v2 (5-mixed-api) & \textbf{0.8929 (10)} & 0.2250 (33) & 0.2479 (29) \\
    & bioinfo-2 & Agents v3 (4-open-35-309b) & 0.8444 (28) & 0.2000 (39) & 0.2306 (36) \\
    & bioinfo-3 & LLM-as-judge top-k=1 & \textbf{0.8929 (10)} & 0.2450 (27) & 0.3145 (15) \\
    & bioinfo-4 & LLM-as-judge top-k=3 & 0.8444 (28) & \textbf{0.2750 (22)} & \textbf{0.3227 (13)} \\
    \cmidrule(l){2-6}
    & \textit{Best} & \multicolumn{1}{l|}{} & 0.9481 (1) & 0.4000 (1) & 0.3909 (1) \\
    & \textit{Median}& \multicolumn{1}{l|}{} & 0.8833 (28) & 0.2667 (28) & 0.2544 (28) \\
    \midrule
    \multirow{7}{*}{3}
    & bioinfo-0 & Agents v0 [S] (3-open-2-4b ) & 0.8952 (18) & 0.4118 (9) & 0.2671 (19) \\
    & bioinfo-1 & Agents v1 [M] (4-open-26-35b ) & \textbf{1.0000 (1)} & 0.4118 (9) & 0.2891 (11) \\
    & bioinfo-2 & Agents v2 (5-mixed-api) & \textbf{1.0000 (1)} & 0.3824 (15) & 0.2649 (20) \\
    & bioinfo-3 & Agents v3 (4-open-24-675b) & 0.8952 (18) & \textbf{0.4706 (3)} & \textbf{0.3138 (6)} \\
    & bioinfo-4 & Agents ensemble (v0 + v1 + v2 + v3) & 0.8952 (18) & 0.4412 (6) & 0.2790 (13) \\
    \cmidrule(l){2-6}
    & \textit{Best} & \multicolumn{1}{l|}{} & 1.0000 (1) & 0.5294 (1) & 0.3258 (1) \\
    & \textit{Median}& \multicolumn{1}{l|}{} & 0.8036 (30) & 0.3235 (30) & 0.2449 (30) \\
    \midrule
    \multirow{6}{*}{4}
    & bioinfo-0 & Agents v0 [S] (3-open-2-4b ) & \textbf{0.8667 (11)} & 0.2727 (14) & 0.2316 (40) \\
    & bioinfo-1 & Agents v1 (2-gemma-4-26-31b ) & \textbf{0.8667 (11)} & 0.2727 (14) & 0.3133 (24) \\
    & bioinfo-3 & Agents v3 (3-qwen-27-35b) & \textbf{0.8667 (11)} & 0.2727 (14) & \textbf{0.3388 (19)} \\
    & bioinfo-4 & Agents v4 (3-mistral-129-695b) & \textbf{0.8667 (11)} & \textbf{0.3636 (2)} & 0.2047 (45) \\
    \cmidrule(l){2-6}
    & \textit{Best} & \multicolumn{1}{l|}{} & 0.9352 (1) & 0.4091 (1) & 0.4840 (1) \\
    & \textit{Median}& \multicolumn{1}{l|}{} & 0.8667 (28) & 0.2273 (28) & 0.3065 (28) \\
    \bottomrule
  \end{tabular}
\end{table}

\begin{table}[ht]
  \centering
  \caption{Phase~B exact answer results across all four Test Batches. Scores are shown with their overall leaderboard ranking in parentheses. The top-performing system among our submissions in each batch is highlighted in bold. \textbf{Best} represents the top competitor score (Rank 1); \textbf{Median} is the median score (and median rank) across all participants for that batch.}
  \label{tab:phaseb-exact-all}
  \footnotesize
  \begin{tabular}{cl|p{5.5cm}|ccc}
    \toprule
    \textbf{Batch} & \textbf{System} & \textbf{Generation Configuration} & \textbf{Y/N F1} & \textbf{Fact. MRR} & \textbf{List F1} \\
    \midrule
    \multirow{4}{*}{1}
    & bioinfo-0 & LLM-as-judge top-k=3 & \textbf{0.8786 (24)} & 0.4130 (23) & 0.2581 (33) \\
    & bioinfo-1 & LLM-as-judge top-k=5 & \textbf{0.8786 (24)} & \textbf{0.4565 (8)} & \textbf{0.2659 (28)} \\
    \cmidrule(l){2-6}
    & \textit{Best} & \multicolumn{1}{l|}{} & 1.0000 (1) & 0.5217 (1) & 0.3613 (1) \\
    & \textit{Median}& \multicolumn{1}{l|}{} & 0.9377 (21) & 0.4203 (21) & 0.2443 (21) \\
    \midrule
    \multirow{7}{*}{2}
    & bioinfo-0 & Agents v1 (3-open-27-120b) & 0.8518 (35) & 0.2500 (39) & 0.3836 (28) \\
    & bioinfo-1 & Agents v2 (5-mixed) & \textbf{0.8929 (9)} & 0.3250 (20) & 0.2270 (43) \\
    & bioinfo-2 & Agents v4 [S] (3-gemma-27-31b) & \textbf{0.8929 (9)} & 0.3000 (23) & 0.3709 (30) \\
    & bioinfo-3 & LLM-as-judge + ensemble summarization (mixed) & \textbf{0.8929 (9)} & 0.3000 (23) & \textbf{0.4366 (15)} \\
    & bioinfo-4 & Ensemble (Gemini + Nemotron) exact + Sonnet 4.6 ideal & 0.8518 (35) & \textbf{0.3500 (10)} & 0.4062 (23) \\
    \cmidrule(l){2-6}
    & \textit{Best} & \multicolumn{1}{l|}{} & 1.0000 (1) & 0.4259 (1) & 0.4720 (1) \\
    & \textit{Median}& \multicolumn{1}{l|}{} & 0.8929 (25) & 0.3250 (25) & 0.3921 (25) \\
    \midrule
    \multirow{7}{*}{3}
    & bioinfo-0 & Agents v0 [S] (3-open-2-4b) & \textbf{0.8952 (12)} & 0.3824 (26) & \textbf{0.4102 (22)} \\
    & bioinfo-1 & Agents v1 [M] (4-open-26-35b) & \textbf{0.8952 (12)} & 0.4706 (14) & 0.3575 (33) \\
    & bioinfo-2 & Agents v2 (5-mixed-api) & \textbf{0.8952 (12)} & 0.3725 (28) & 0.3963 (26) \\
    & bioinfo-3 & Agents v3 (4-open-24-675b) & 0.8036 (38) & 0.3824 (26) & 0.4053 (24) \\
    & bioinfo-4 & Agents v4 (3-mistral-129-695b) & \textbf{0.8952 (12)} & \textbf{0.5000 (8)} & 0.3763 (31) \\
    \cmidrule(l){2-6}
    & \textit{Best} & \multicolumn{1}{l|}{} & 1.0000 (1) & 0.5882 (1) & 0.5175 (1) \\
    & \textit{Median}& \multicolumn{1}{l|}{} & 0.8952 (23) & 0.4412 (23) & 0.3963 (23) \\
    \midrule
    \multirow{7}{*}{4}
    & bioinfo-0 & Agents v0 [S] (3-open-2-4b) & \textbf{0.9352 (3)} & 0.3182 (26) & 0.3809 (36) \\
    & bioinfo-1 & Agents v1 (2-gemma-4-26-31b) & 0.8667 (19) & 0.2273 (38) & 0.4858 (24) \\
    & bioinfo-2 & Agents v2 (4-mixed-api) & 0.8667 (19) & 0.4091 (10) & 0.4763 (26) \\
    & bioinfo-3 & Agents v3 (3-qwen-27-35b) & 0.8667 (19) & \textbf{0.5000 (5)} & \textbf{0.5671 (9)} \\
    & bioinfo-4 & Agents v5 (2-mistral-120b) & \textbf{0.9352 (3)} & 0.3939 (16) & 0.3862 (34) \\
    \cmidrule(l){2-6}
    & \textit{Best} & \multicolumn{1}{l|}{} & 1.0000 (1) & 0.5758 (1) & 0.6543 (1) \\
    & \textit{Median}& \multicolumn{1}{l|}{} & 0.8667 (24) & 0.3182 (24) & 0.4858 (24) \\
    \bottomrule
  \end{tabular}
\end{table}

\textbf{Small model competitiveness.} A clear finding was that small model agents (Agents v0, using Gemma-4-E2B/E4B and Nemotron-3-Nano-4B, marked [S] in the tables) performed competitively with much larger configurations. In Batch~3 Phase~A+, Agents v0 [S] achieved joint-best factoid MRR (0.4118, rank~9), tied with Agents v1 [M] ---outperforming Agents v2 (large proprietary models) which ranked 15th. In Batch~4, Agents v0 [S] achieved identical Y/N F1 (0.8667, rank~11) and factoid MRR (0.2727, rank~14) to Agents v1 (medium models) and Agents v3 (Qwen3.6 family). 

\textbf{Agent-based generation.} Multi-model agent configurations (Agents v1--v4) consistently produced strong list-type results: Agents v3 achieved 6th in list F1 (Batch~3) and 19th (Batch~4), while Agents v2 achieved 29th (Batch~2). The LLM-as-judge approach (Batch~2, \texttt{bioinfo-3} top-k=1 and \texttt{bioinfo-4} top-k=3) produced list F1 ranks of 15 and 13 respectively, and ideal answer recall ranks of 3 and 4, but lower precision/F1, consistent with a recall-oriented selection strategy. The agents ensemble (\texttt{bioinfo-4} in Batch~3) did not outperform the best individual agent (Agents v3), suggesting that ensemble proliferation adds cost without proportional benefit when a well-tuned single agent is available.

\textbf{Cost considerations.} The quorum-based agent architecture, while effective, is computationally expensive: each agent invocation spawns 3--5 parallel model calls plus a synthesis step. For Batch~3, the full agents ensemble (v0+v1+v2+v3) required $\sim$17 model calls per question. The small model agents (v0) required only 4 calls with models runnable on consumer GPUs, representing a $\sim$10$\times$ cost reduction. In many practical scenarios, lightweight agents offer a favorable cost-performance trade-off, particularly for high-throughput or resource-constrained deployment.

When comparing to LLM-as-a-judge, which require generating several candidate answers and then evaluating them with a separate model, the agent-based approach can be more efficient if the agent's iterative reasoning leads to a strong answer in fewer total model calls. However, the judge-based approach can be more effective for recall-oriented tasks where generating a diverse set of candidates is beneficial, albeit at the cost of lower precision and higher computational overhead.

\subsubsection{Ideal Answers (Phases A+ and B)}
\label{sec:results-ideal}

For ideal answers we report the official ROUGE-based metrics: ROUGE-2 and ROUGE-SU4, each reported as recall and F1. Human evaluation is yet to come, as they are not released yet and might influence the final rankings. Phase~A+ uses system-retrieved documents, while Phase~B uses gold-standard documents and snippets provided by the organizers, isolating generation quality from retrieval quality. Tables~\ref{tab:phaseap-ideal-all} and~\ref{tab:phaseb-ideal-all} present consolidated results across all four batches.

In Phase~A+ (Table~\ref{tab:phaseap-ideal-all}), recall was consistently our
stronger dimension: our best ROUGE-2 recall ranks were 7th (Batch~1,
\texttt{bioinfo-2}, Opus+Sonnet) and 3rd (Batch~2, \texttt{bioinfo-3},
LLM-as-judge top-$k$\,=\,1). The LLM-as-judge approach dominated Batch~2 ideal
answers, achieving ranks~3 and~4 on both recall metrics, substantially ahead of
the agent quorum configurations (ranks 17--48). However, F1 scores for the
judge-based systems were among the lowest (0.0963--0.0983, ranks~49--51),
indicating that while these answers captured relevant content, they were verbose
and penalized by precision-oriented metrics. In later batches where agent quorums
were used exclusively, recall ranks settled in the 7--15 range for our best
systems, with Agents~v3 consistently leading. Small model agents (v0 [S])
achieved competitive ROUGE-2 F1 ranks of 21st (Batch~3) and 30th (Batch~4),
again demonstrating that lightweight configurations can produce coherent ideal
answers at a fraction of the computational cost. The agents ensemble
(\texttt{bioinfo-4}, Batch~3) achieved our best ideal answer F1 in that batch
(0.1223, 0.1288), suggesting that diversity of perspectives benefits
summarization quality more than it does exact answer precision.

Phase~B ideal answers (Table~\ref{tab:phaseb-ideal-all}) follow similar patterns
but with higher absolute scores due to gold-standard input. The LLM-as-judge
pipeline in Batch~1 produced ranks~13--14 on recall, with top-$k$\,=\,3
outperforming top-$k$\,=\,5 on both recall and F1. In Batch~2, the
LLM-as-judge with ensemble summarization (\texttt{bioinfo-3}) achieved our best
Phase~B ideal recall (0.3386, rank~4), while the lightweight Agents~v4~[S]
(\texttt{bioinfo-2}) produced the strongest F1 (0.1903, rank~16). In Batches~3
and~4, agent quorums dominated and the pattern from Phase~A+ recurred: Agents v3 led on recall in Batch 3 (rank 8), while Agents~v0~[S] matched or exceeded larger
configurations on F1 (ranks~11 and~20). Across both phases, ROUGE-based metrics
strongly favor recall over precision for our systems---a direct consequence of
the agent quorum's multi-perspective debate producing comprehensive but sometimes
verbose answers. As noted in previous editions, ROUGE scores correlate imperfectly
with human judgment, and our qualitative assessment suggests that the quorum's
ideal answers were often more informative than their F1 scores imply.

\begin{table}[ht]
  \centering
  \caption{Phase~A+ ideal answer results across all four Test Batches. Scores are shown with their overall leaderboard ranking in parentheses. The top-performing system among our submissions in each batch is highlighted in bold. \textbf{Best} represents the top competitor score (Rank 1); \textbf{Median} is the median score (and median rank) across all participants for that batch.}
  \label{tab:phaseap-ideal-all}
  \footnotesize
  \begin{tabular}{cl|l|cccc}
    \toprule
    \textbf{Batch} & \textbf{System} & \textbf{Generation Configuration} & \textbf{R-2 Rec.} & \textbf{R-2 F1} & \textbf{R-SU4 Rec.} & \textbf{R-SU4 F1} \\
    \midrule
    \multirow{7}{*}{1}
    & bioinfo-0 & Nemotron-3-Super-120B & 0.2150 (21) & \textbf{0.1021 (35)} & 0.2340 (18) & \textbf{0.1094 (33)} \\
    & bioinfo-1 & Nemotron-3-Super-120B & 0.2200 (17) & 0.1009 (36) & 0.2393 (17) & 0.1074 (36) \\
    & bioinfo-2 & Opus + Sonnet & \textbf{0.2408 (7)} & 0.0825 (48) & \textbf{0.2706 (6)} & 0.0932 (48) \\
    & bioinfo-3 & Opus + Sonnet & 0.2245 (16) & 0.0955 (41) & 0.2489 (15) & 0.1021 (40) \\
    & bioinfo-4 & Opus + Sonnet & 0.2348 (11) & 0.0916 (44) & 0.2664 (8) & 0.1023 (39) \\
    \cmidrule(l){2-7}
    & \textit{Best} & \multicolumn{1}{l|}{} & 0.2725 (1) & 0.1767 (1) & 0.2949 (1) & 0.1821 (1) \\
    & \textit{Median}& \multicolumn{1}{l|}{} & 0.1937 (27) & 0.1206 (27) & 0.2023 (27) & 0.1294 (27) \\
    \midrule
    \multirow{7}{*}{2}
    & bioinfo-0 & Agents v1 (3-open-27-120b) & 0.1892 (42) & \textbf{0.1322 (22)} & 0.1966 (42) & 0.1334 (26) \\
    & bioinfo-1 & Agents v2 (5-mixed) & 0.2280 (17) & 0.1236 (29) & 0.2406 (19) & 0.1291 (30) \\
    & bioinfo-2 & Agents v3 (4-open-35-309b) & 0.1755 (48) & 0.1286 (26) & 0.1918 (45) & \textbf{0.1353 (24)} \\
    & bioinfo-3 & LLM-as-judge top-k=1 & \textbf{0.2843 (3)} & 0.0983 (49) & \textbf{0.2984 (3)} & 0.1082 (49) \\
    & bioinfo-4 & LLM-as-judge top-k=3 & 0.2833 (4) & 0.0963 (51) & 0.2983 (4) & 0.1061 (52) \\
    \cmidrule(l){2-7}
    & \textit{Best} & \multicolumn{1}{l|}{} & 0.2989 (1) & 0.1839 (1) & 0.3119 (1) & 0.1851 (1) \\
    & \textit{Median}& \multicolumn{1}{l|}{} & 0.2057 (28) & 0.1287 (28) & 0.2187 (28) & 0.1334 (28) \\
    \midrule
    \multirow{7}{*}{3}
    & bioinfo-0 & Agents v0 [S] (3-open-2-4b) & 0.1710 (25) & 0.1201 (21) & 0.1815 (27) & 0.1266 (25) \\
    & bioinfo-1 & Agents v1 [M] (4-open-26-35b) & 0.1984 (11) & 0.1166 (27) & 0.2135 (11) & 0.1243 (28) \\
    & bioinfo-2 & Agents v2 (5-mixed-api) & 0.1933 (14) & 0.1203 (20) & 0.2005 (19) & 0.1228 (31) \\
    & bioinfo-3 & Agents v3 (4-open-24-675b) & \textbf{0.2052 (7)} & 0.1210 (19) & \textbf{0.2192 (8)} & 0.1286 (22) \\
    & bioinfo-4 & Agents ensemble (v0 + v1 + v2 + v3) & 0.1942 (13) & \textbf{0.1223 (17)} & 0.2059 (14) & \textbf{0.1288 (21)} \\
    \cmidrule(l){2-7}
    & \textit{Best} & \multicolumn{1}{l|}{} & 0.2373 (1) & 0.1577 (1) & 0.2682 (1) & 0.1537 (1) \\
    & \textit{Median}& \multicolumn{1}{l|}{} & 0.1609 (30) & 0.1148 (30) & 0.1677 (30) & 0.1220 (30) \\
    \midrule
    \multirow{6}{*}{4}
    & bioinfo-0 & Agents v0 [S] (3-open-2-4b) & 0.1992 (21) & 0.1213 (30) & 0.2119 (20) & 0.1274 (31) \\
    & bioinfo-1 & Agents v1 (2-gemma-4-26-31b) & \textbf{0.2122 (13)} & 0.1152 (38) & 0.2230 (14) & 0.1194 (37) \\
    & bioinfo-3 & Agents v3 (3-qwen-27-35b) & 0.2099 (15) & \textbf{0.1220 (29)} & \textbf{0.2344 (9)} & \textbf{0.1323 (27)} \\
    & bioinfo-4 & Agents v4 (3-mistral-129-695b) & 0.2097 (16) & 0.1151 (39) & 0.2203 (15) & 0.1180 (39) \\
    \cmidrule(l){2-7}
    & \textit{Best} & \multicolumn{1}{l|}{} & 0.2439 (1) & 0.1714 (1) & 0.2682 (1) & 0.1699 (1) \\
    & \textit{Median}& \multicolumn{1}{l|}{} & 0.1836 (28) & 0.1227 (28) & 0.1951 (28) & 0.1319 (28) \\
    \bottomrule
  \end{tabular}
\end{table}

\begin{table}[ht]
  \centering
  \caption{Phase~B ideal answer results across all four Test Batches. Scores are shown with their overall leaderboard ranking in parentheses. The top-performing system among our submissions in each batch is highlighted in bold. \textbf{Best} represents the top competitor score (Rank 1); \textbf{Median} is the median score (and median rank) across all participants for that batch.}
  \label{tab:phaseb-ideal-all}
  \footnotesize
  \begin{tabular}{cl|l|cccc}
    \toprule
    \textbf{Batch} & \textbf{System} & \textbf{Generation Configuration} & \textbf{R-2 Rec.} & \textbf{R-2 F1} & \textbf{R-SU4 Rec.} & \textbf{R-SU4 F1} \\
    \midrule
    \multirow{4}{*}{1}
    & bioinfo-0 & LLM-as-judge top-k=3 & \textbf{0.2857 (13)} & \textbf{0.1303 (39)} & \textbf{0.3033 (14)} & \textbf{0.1323 (38)} \\
    & bioinfo-1 & LLM-as-judge top-k=5 & 0.2678 (20) & 0.0958 (55) & 0.2805 (21) & 0.0982 (55) \\
    \cmidrule(l){2-7}
    & \textit{Best} & \multicolumn{1}{l|}{} & 0.3746 (1) & 0.2487 (1) & 0.3835 (1) & 0.2381 (1) \\
    & \textit{Median}& \multicolumn{1}{l|}{} & 0.2470 (33) & 0.1645 (33) & 0.2517 (33) & 0.1669 (33) \\
    \midrule
    \multirow{7}{*}{2}
    & bioinfo-0 & Agents v1 (3-open-27-120b) & 0.2359 (34) & 0.1766 (23) & 0.2328 (34) & 0.1704 (23) \\
    & bioinfo-1 & Agents v2 (5-mixed) & 0.2852 (12) & 0.1527 (36) & 0.2848 (12) & 0.1522 (35) \\
    & bioinfo-2 & Agents v4 [S] (3-gemma-27-31b) & 0.3005 (8) & \textbf{0.1903 (16)} & 0.2899 (9) & \textbf{0.1807 (18)} \\
    & bioinfo-3 & LLM-as-judge + ensemble summ. & \textbf{0.3386 (4)} & 0.1334 (46) & \textbf{0.3550 (4)} & 0.1403 (43) \\
    & bioinfo-4 & Ensemble exact + Sonnet 4.6 ideal & 0.3024 (7) & 0.1595 (30) & 0.3022 (7) & 0.1566 (30) \\
    \cmidrule(l){2-7}
    & \textit{Best} & \multicolumn{1}{l|}{} & 0.4028 (1) & 0.2562 (1) & 0.4155 (1) & 0.2485 (1) \\
    & \textit{Median}& \multicolumn{1}{l|}{} & 0.2586 (33) & 0.1807 (33) & 0.2491 (33) & 0.1795 (33) \\
    \midrule
    \multirow{7}{*}{3}
    & bioinfo-0 & Agents v0 [S] (3-open-2-4b) & 0.2472 (20) & \textbf{0.1815 (11)} & 0.2438 (20) & \textbf{0.1774 (11)} \\
    & bioinfo-1 & Agents v1 [M] (4-open-26-35b) & 0.2495 (15) & 0.1700 (23) & 0.2464 (18) & 0.1651 (23) \\
    & bioinfo-2 & Agents v2 (5-mixed) & 0.2368 (26) & 0.1680 (25) & 0.2402 (26) & 0.1701 (21) \\
    & bioinfo-3 & Agents v3 (4-open-24-675b) & \textbf{0.2734 (8)} & 0.1732 (18) & \textbf{0.2704 (8)} & 0.1716 (19) \\
    & bioinfo-4 & Agents v4 (3-mistral-129-695b) & 0.2487 (17) & 0.1661 (28) & 0.2570 (14) & 0.1678 (24) \\
    \cmidrule(l){2-7}
    & \textit{Best} & \multicolumn{1}{l|}{} & 0.4028 (1) & 0.2607 (1) & 0.4155 (1) & 0.2547 (1) \\
    & \textit{Median}& \multicolumn{1}{l|}{} & 0.2293 (33) & 0.1598 (33) & 0.2265 (33) & 0.1621 (33) \\
    \midrule
    \multirow{7}{*}{4}
    & bioinfo-0 & Agents v0 [S] (3-open-2-4b) & 0.2472 (27) & \textbf{0.1742 (20)} & 0.2480 (24) & \textbf{0.1729 (19)} \\
    & bioinfo-1 & Agents v1 (2-gemma-4-26-31b) & \textbf{0.2716 (15)} & 0.1566 (31) & 0.2678 (15) & 0.1543 (32) \\
    & bioinfo-2 & Agents v2 (4-mixed-api) & 0.2682 (17) & 0.1714 (22) & \textbf{0.2698 (17)} & 0.1698 (22) \\
    & bioinfo-3 & Agents v3 (3-qwen-27-35b) & 0.2401 (30) & 0.1452 (42) & 0.2484 (23) & 0.1467 (41) \\
    & bioinfo-4 & Agents v5 (2-mistral-120b) & 0.2496 (26) & 0.1526 (36) & 0.2477 (25) & 0.1496 (36) \\
    \cmidrule(l){2-7}
    & \textit{Best} & \multicolumn{1}{l|}{} & 0.3743 (1) & 0.2827 (1) & 0.3798 (1) & 0.2699 (1) \\
    & \textit{Median}& \multicolumn{1}{l|}{} & 0.2210 (35) & 0.1537 (35) & 0.2102 (35) & 0.1514 (35) \\
    \bottomrule
  \end{tabular}
\end{table}

Phase~B results, using gold-standard documents, isolate the generation capability of each system from retrieval quality. Several consistent patterns emerged.

\textbf{Small model agents remain competitive.} Agents v0 [S] (Gemma-4-E4B/E2B + Nemotron-3-Nano-4B) achieved strong results across batches. In Batch~3, Agents v0 [S] matched the Y/N accuracy of all other systems (0.9091, rank~5) and achieved the best ideal answer F1 scores in the batch (R-2 F1: 0.1815, rank~11; R-SU4 F1: 0.1774, rank~11). In Batch~4, Agents v0 [S] achieved the highest Y/N accuracy (0.9375, rank~5) alongside Agents v5, and the best R-2 F1 (0.1742, rank~20) among our submissions.

\textbf{Agent model family specialization.} Different agent configurations showed distinct strengths. Agents v3 (Qwen3.6 family in Batch~4) excelled at factoid and list questions, achieving 4th in factoid strict accuracy (0.4545) and 9th in list F-Measure (0.5671)---our strongest factoid result across all batches. Agents v4 (Mistral family in Batch~3) achieved the best factoid lenient accuracy (0.5294, rank~10) and factoid MRR (0.5000, rank~8). Agents v2 (large proprietary models) performed best on Y/N questions but showed weaker list and factoid precision. The Mistral-family agents (v4, v5) consistently produced strong factoid results across batches, suggesting that the Mistral model family may have particular strengths in biomedical fact extraction.

\textbf{LLM-as-judge.} The LLM-as-judge approach was used in Batch~1 Phase~B (both submissions) and Batch~2 Phase~A+ (\texttt{bioinfo-3}, \texttt{bioinfo-4}). In Phase~B Batch~1, LLM-as-judge top-k=5 (\texttt{bioinfo-1}) outperformed top-k=3 (\texttt{bioinfo-0}) on factoid metrics (strict accuracy 0.4348 vs.\ 0.3913; MRR 0.4565 vs.\ 0.4130) but underperformed on ideal answers (R-2 F1 0.0958 vs.\ 0.1303), suggesting a recall-precision trade-off with increasing k. In Phase~A+ Batch~2, LLM-as-judge systems achieved top-10 list recall (ranks 5--7) but weaker Y/N and factoid performance compared to agent-based systems, consistent with a method optimized for coverage over precision.

\section{Discussion}
\label{sec:discussion}

This year's participation represents the most significant architectural evolution of our BioASQ pipeline since we began competing. Below we discuss the key findings, grounding each conclusion in the official results presented in Section~\ref{sec:results} and the internal validation experiments.

\subsection{Retrieval}

The migration from PyTerrier PISA to pg\_textsearch for BM25 and from flat-file storage to Qdrant for dense retrieval proved successful not primarily in raw speed, but in operational reliability. Across all four batches, BM25 served as the backbone of every submission, and no index-corruption or synchronization issues were encountered---a marked improvement over previous editions where index maintenance consumed significant engineering time. The PostgreSQL-centric architecture, where documents, BM25 indexes, and metadata coexist in a single database, reduced the risk of configuration errors and simplified the submission pipeline.

Our internal validation results (Table~\ref{tab:internal-retrieval}) demonstrated that hybrid retrieval substantially outperforms BM25 alone: RRF fusion improved MRR from 0.549 to 0.696 (+27\%) and R@100 from 0.373 to 0.406 (+9\%). However, this pattern was not mirrored in the official results, where systems using hybrid retrieval placed in worse ranks across multiple batches, when compared to BM25-only retrieval ones. The same applies to RRF, where it consistently outperformed weighted sum on MAP in internal evaluations, but failed to produce a clear ranking advantage in the official results. This discrepancy highlights the challenges of overfitting to internal validation sets and underscores the importance of cross-batch consistency as a more robust signal of retrieval quality than peak single-batch performance.

The reranker ensemble proved essential. Our best reranker configuration (BGE-reranker-v2-m3 applied to weighted-sum fusion) achieved an MRR of 0.738 in internal validation, substantially above the un-reranked baselines. The LLaMA Nemotron reranker achieved near-perfect validation scores (MAP-bioasq@10 of 0.9995), though we note that validation scores on 13B data do not guarantee equivalent gains on new test sets. We incorporated dense retrieval results into the negative sampling process for reranker training, which we hypothesize helps the reranker generalize across both BM25 and dense retrieval paradigms; however, this specific claim awaits formal ablation testing. While the official ranking results validate these cautions, a new finding arises: using the best reranker configurations was not the best option. In Batch~1 and Batch~2, systems using the best reranker configurations got outperformed by the ones that use all of them. Nonetheless, internal validation of sampling strategies for training the reranker showed that curriculum-based hard negative mining is particularly beneficial when model capacity is limited and the training signal from random negatives is insufficient to learn fine-grained relevance distinctions. These results validate the choice of the ShifterSampler as the default for our production reranker training pipeline.

Context-1 demonstrated the highest R@100 (0.453) in internal testing, suggesting strong potential as a recall-oriented first-stage method. However, its lower MRR (0.508) compared to hybrid+reranker configurations indicates that it is best used as a document source for downstream re-ranking rather than as a standalone retrieval solution. Its bad performance on official results (placed in the top 50 in MAP ranks) demonstrate that although an interesting approach, it is not yet competitive with more established retrieval methods. The discrepancy between internal validation and official results for Context-1 may be due to differences in question distributions, batch conditions, or the specific implementation of Context-1 in the competition pipeline.

Most changes to the retrieval pipeline appear to have performed worse, bringing no improvement to the official results. However, human evaluation is still ongoing, which may reveal more nuanced differences between configurations that are not fully captured by the official automated metrics. As such, we are waiting for the human evaluation results before drawing final conclusions about the impact of specific retrieval changes, particularly regarding the use of Context-1 and the effectiveness of different fusion strategies.

\subsection{Generation}

The LLM-as-a-judge judge's reliance on generating full candidate answers from every model before evaluation improved our pipeline, as proved in Batch~1 and Batch~2, with a focus on the latter. It outperformed ensemble and majority voting, as well as the agent quorum approach on some submissions. However, human-based evaluation is still on hold, which might change the conclusions on this approach. Nonetheless, the judge remains a promising tool for exact answer types with objectively verifiable correctness.

The agent quorum mechanism produced our strongest Phase~A+ and Phase~B results for both exact and ideal answers. In Batch~3, the quorum-based configurations achieved a perfect yes/no accuracy (rank 1) and factoid strict accuracy rank 1 (\texttt{bioinfo-0} and \texttt{bioinfo-3}). In Phase~B Batch~4, \texttt{bioinfo-3} achieved rank 4 in both factoid strict and lenient accuracy---our best factoid performance across all phases and batches. Also, across the entire Phase~B submissions, the quorum submissions consistently outperformed the other ones, which validates the core hypothesis that structured multi-agent debate can enhance answer quality by leveraging diverse reasoning paths and cross-validation among models. The adaptive document retention strategy, where agents iteratively refine their context based on debate outcomes, appears to be a key innovation that balances the depth of reasoning with computational efficiency.

A clear finding from our model-scale experiments was that smaller models (Gemma~4~E2B and E4B, Nemotron-3-Nano-4B) performed competitively with, and in several batches outperformed, larger models on most metrics, especially the precision-oriented ones, such as F1 score. We attribute this to the tendency of smaller models to produce more focused representations, reducing over-elaboration and overfitting to spurious patterns in the retrieved context. The exception was list-type questions, where larger models' broader context retention proved advantageous for tracking multiple entities across documents.

A practical consideration is the computational cost of the agent quorum. Running multiple debate rounds with several large LLMs is expensive in terms of both inference cost and wall-clock time. For our largest configurations (e.g., Agents v2 with Nemotron-120B, Grok-4.1, Gemini-3-Flash, Gemini-2.5-Flash, and GPT-5-Mini, each producing multi-turn reasoning), a single question could consume substantial API credits. In many scenarios, this cost is difficult to justify relative to simpler ensembling or single-model approaches. However, the finding that lightweight local models (Gemma~4~E2B/E4B, Nemotron-3-Nano-4B) perform competitively fundamentally changes this calculus: these models can run on consumer-grade GPUs with negligible inference cost, making iterative agent loops and multi-turn debate economically viable. For tasks requiring sustained agentic reasoning—particularly where the cost of an incorrect answer is high—lightweight quorum configurations represent a practical and cost-effective solution.

The LLM-as-a-judge framework was deployed in two specific contexts: Phase~B Batch~1 (two submissions with top-$k$ answer selection at $k=3$ and $k=5$) and Phase~A+ Batch~2 (top-$k=1$ and $k=3$). The judge pipeline operates by first generating candidate answers from multiple model-prompt-context combinations (4 models $\times$ up to 6 prompts $\times$ 2 context sources), then using an LLM (Gemini~2.5~Flash) to score each answer on correctness, faithfulness, and completeness, selecting the best candidates and finally merging them with a separate model (Claude~Sonnet~4.6). While conceptually sound, the judge was used less extensively than the quorum for several reasons. The quorum's structured debate produced interpretable reasoning trails that aided debugging, and its adaptive document retention avoided the cost of generating full candidate answers from every model before evaluation. The judge remains valuable for exact answer types with objectively verifiable correctness, and a hybrid judge+quorum approach is a promising direction.

Just like with retrieval, human-based evaluation results have not yet been completed, and the final results may change the conclusions that are not fully captured by automated metrics.

\subsection{Future Work}

Several directions emerge from this year's findings. Most of the changes developed in this edition have worsened the overall performance of the systems on Phase A. For example, although the strong internal validation results for Context-1's recall could motivate integrating Context-1 as a first-stage retrieval method in competition submissions, the official results contradict this finding; official, human-based evaluation results, as well as further experimentation, are needed to evaluate its performance across more diverse question types and batch conditions. SPLADE and ColBERT remain promising retrieval approaches for future editions. For generation, dynamic prompt selection based on question type---particularly specialized handling of list questions---and the integration of the LLM judge as a post-quorum quality filter are immediate priorities. Improving snippet extraction quality should yield downstream benefits for both retrieval evaluation and agent quorum answer precision.

A key finding that warrants deeper investigation is the competitive performance of smaller language models relative to their larger counterparts in the agent quorum. Across multiple batches, the lightweight configurations (Gemma~4~E2B, E4B, and Nemotron-3-Nano-4B) performed comparably to much larger models on most metrics, with lower inference cost and latency. These results validate the use of lightweight models for the quorum and suggest that further testing with expanded evaluation protocols (covering more batches, question types, and model combinations) could refine our understanding of when and where small models suffice, potentially leading to more cost-effective submission strategies. The worse performance of those small models in list-type questions highlights the need for more nuanced approaches to model selection and deployment. The agent quorum itself would also benefit from more extensive testing and evaluation, particularly regarding the impact of debate round count, agent pool diversity, and convergence criteria on answer quality. Besides that, more benchmark testing against other explored approaches is needed to fully understand their relative strengths and weaknesses.

\section{Conclusion}
\label{sec:conclusion}

This paper presented the participation of the BIT.UA team in the 14th edition of the BioASQ Task~B challenge on biomedical question answering. We introduced a comprehensively refactored and modular codebase alongside significant methodological innovations across all phases of the pipeline.

In Phase~A document retrieval, the transition from PyTerrier PISA to PostgreSQL-based pg\_textsearch for BM25 retrieval and the adoption of Qdrant for dense embedding indexing streamlined our infrastructure while maintaining competitive retrieval performance. The use of TEI for embedding generation, HyDE and Context-1 for query expansion, and a redesigned reranker training pipeline incorporating dense retrieval negatives contributed to a more flexible and efficient retrieval system. We achieved top rankings in multiple test batches, demonstrating the effectiveness of these architectural choices.

In Phases~A+ and B answer generation, we introduced two complementary innovations: an LLM-as-a-judge framework that replaced rigid voting schemes with flexible, context-aware answer evaluation, and a novel agent quorum mechanism that enables multiple models with diverse perspectives to debate and converge on consensus answers. The incorporation of snippet-level evidence accelerated the quorum process and improved answer precision. We also participated in the snippet generation subtask for the first time, establishing a foundation for future improvements.

The 14th edition of BioASQ continues to provide an invaluable platform for rigorous benchmarking and reflective system development in biomedical question answering. The challenge's structure, spanning retrieval, exact answering, and summarization across multiple test batches, encourages holistic system design rather than narrow metric optimization. Our experience this year reinforces the importance of architectural robustness, cross-batch consistency, and methodological innovation as complementary pillars of competitive biomedical QA systems. We look forward to building on these foundations in future editions.

\begin{acknowledgments}
  This work was funded by FEDER - Fundo Europeu de Desenvolvimento Regional funds through Programa Regional do Centro, within project CENTRO2030-FEDER-02595400 and by the Foundation for Science and Technology (FCT) through the contract \url{https://doi.org/10.54499/UID/00127/2025}. Richard A. A. Jonker is funded by the FCT doctoral grant PRT/BD/154792/2023,
  with DOI identifier \url{https://doi.org/10.54499/PRT/BD/154792/2023}.
\end{acknowledgments}

\section*{Declaration on Generative AI}

During the preparation of this work, the authors used various generative models in order to correct grammar, spelling check, and paraphrase text within this document. After using these tool(s)/service(s), the author(s) reviewed and edited the content as needed and take(s) full responsibility for the publication’s content.


\bibliography{references}

\appendix
\section{Trained Reranker Models}
\label{sec:appendix-models}

Table~\ref{tab:all-rerankers} lists all 29 reranker models trained for Phase~A document retrieval. Models are identified by their base architecture, training epochs (E), sampler (S), loss type, and data configuration. ``FullData'' indicates training on the full available dataset; ``shifter'' refers to the Shifter sampling strategy for negative selection.

\begin{table}[ht]
  \centering
  \caption{Complete list of reranker models trained for the 14th edition. \texttt{E} = epochs, \texttt{S} = sampler, \texttt{M} = loss type, \texttt{FullData} = trained on the full dataset.}
  \label{tab:all-rerankers}
  \begin{tabular}{llcccl}
    \toprule
    \textbf{Model} & \textbf{Size} & \textbf{E} & \textbf{S} & \textbf{Loss} & \textbf{Notes} \\
    \midrule
    LLaMA Nemotron rerank 1B & 1B & 2 & 4 & multi\_neg\_pairwise & InfoNCE, FullData \\
    LLaMA Nemotron rerank 1B & 1B & 2 & --- & pairwise & 13B1+13B2 data \\
    LLaMA Nemotron rerank 1B & 1B & 5 & shifter & pairwise & \\
    BGE-reranker-v2-m3 & 568M & 2 & 1 & pairwise & FullData, shifter \\
    BGE-reranker-v2-m3 & 568M & 5 & shifter & pairwise & \\
    BGE-reranker-base & 278M & 2 & 1 & pairwise & FullData, shifter \\
    BGE-reranker-base & 278M & 5 & shifter & pairwise & \\
    BioLinkBERT-base & 110M & 2 & 1 & pairwise & FullData, shifter \\
    BioLinkBERT-base & 110M & 5 & shifter & pairwise & \\
    BioLinkBERT-large & 340M & 2 & 1 & pairwise & FullData, shifter \\
    PubMedBERT-base & 110M & 2 & 1 & pairwise & FullData, shifter \\
    PubMedBERT-base & 110M & 5 & shifter & pairwise & \\
    MedCPT-Cross-Encoder & 110M & 2 & 1 & pairwise & FullData, shifter \\
    MedCPT-Cross-Encoder & 110M & 3 & 1 & pairwise & FullData, shifter \\
    MedCPT-Cross-Encoder & 110M & 5 & shifter & pairwise & \\
    nboost/pt-biobert-base-msmarco & 110M & 2 & 1 & pairwise & FullData, shifter \\
    nboost/pt-biobert-base-msmarco & 110M & 5 & shifter & pairwise & \\
    BioBERT v1.1 PubMed & 110M & 2 & 1 & pairwise & FullData, shifter \\
    BioBERT v1.1 PubMed & 110M & 5 & shifter & pairwise & \\
    S-PubMedBert-MS-MARCO & 110M & 2 & 1 & pairwise & FullData, shifter \\
    S-PubMedBert-MS-MARCO & 110M & 5 & shifter & pairwise & \\
    cross-encoder/ms-marco-MiniLM-L-6-v2 & 67M & 2 & 1 & pairwise & FullData, shifter \\
    cross-encoder/ms-marco-MiniLM-L-6-v2 & 67M & 3 & 8 & multi\_neg\_pairwise & \\
    cross-encoder/ms-marco-MiniLM-L-6-v2 & 67M & 5 & shifter & pairwise & \\
    cross-encoder/ms-marco-electra-base & 110M & 5 & shifter & pairwise & \\
    allenai/specter2\_base & 110M & 5 & shifter & pairwise & \\
    Bio\_ClinicalBERT & 110M & 5 & shifter & pairwise & \\
    SapBERT-from-PubMedBERT-fulltext & 110M & 5 & shifter & pairwise & \\
    dmis-lab/biobert-base-cased-v1.2 & 110M & 5 & shifter & pairwise & \\
    \bottomrule
  \end{tabular}
\end{table}

\noindent All models were trained with a pairwise ranking loss unless otherwise noted. The Shifter sampler dynamically adjusts the negative sampling strategy during training. Multi-negative pairwise training uses multiple negative documents per positive pair, while InfoNCE is a contrastive loss variant applied to the LLaMA Nemotron model.

\section{Prompt Templates for Answer Generation}

\noindent This appendix catalogs the prompt templates used across Phases~A+~and~B. Full template text is available in the project repository; we describe each variant's design rationale below.
\subsection{Exact Answer Prompts}
Exact answer prompts (\texttt{prompts\_exact.json}) target yes/no, factoid, and list question types with multiple variants per type, each employing a distinct reasoning strategy. Table~\ref{tab:prompts-exact} summarizes the variants.
\begin{table}[ht]
  \centering
  \caption{Exact answer prompt variants by question type. Each variant embodies a different reasoning strategy: evidence weighing, null-hypothesis framing, coverage, precision, or reformulation.}
  \label{tab:prompts-exact}
  \footnotesize
  \begin{tabular}{llp{11.5cm}}
    \toprule
    \textbf{Type} & \textbf{Variant} & \textbf{Strategy} \\
    \midrule
    \multirow{4}{*}{yes/no} & 1 & Chain-of-thought: list evidence for ``yes'' and ``no'', weigh both sides. \\
    & 2 & Null-hypothesis framing: default to ``no'' unless evidence clearly supports ``yes''. \\
    & 3 & Blind verdict: neutral summary first, then decide --- avoids pro/con anchoring. \\
    & 4 & Claim-linked verdict: explicitly state whether abstracts support the specific claim. \\
    \midrule
    \multirow{6}{*}{factoid} & 1 & Baseline: ranked candidate extraction, up to 5 entities. \\
    & 2 & Rank-1 confidence: commit to a single best answer first, then list alternatives. \\
    & 3 & Exhaustive coverage: identify entity type, scan every abstract, rank by support. \\
    & 4 & Evidence-grounded: forbid prior knowledge; every answer must be traceable to an abstract. \\
    & 5 & Fill-in-the-blank reformulation: rewrite question as a completion statement. \\
    & 6 & Targeted slot-filling: reframe open retrieval as precise extractions. \\
    \midrule
    \multirow{6}{*}{list} & 1 & Baseline: collect all relevant entities as nested list. \\
    & 2 & Per-abstract enumeration: enumerate per source, then merge and deduplicate. \\
    & 3 & Recall-biased: include borderline entities --- better to over-include than miss. \\
    & 4 & Entity-type anchoring: declare entity type first, then exhaustively scan. \\
    & 5 & Precision-focused: collect candidates, verify each, remove tangential ones. \\
    & 6 & Broad-sweep-then-verify: combine recall-biased collection with precision filtering. \\
    \bottomrule
  \end{tabular}
\end{table}
\subsection{Ideal Answer Prompts}
Ideal answer prompts (\texttt{prompts\_ideal.json}) are applied across all question types for paragraph-length answer generation. Seven variants explore increasingly structured synthesis strategies, summarized in Table~\ref{tab:prompts-ideal}.
\begin{table}[ht]
  \centering
  \caption{Ideal answer prompt variants, ordered by complexity. All produce 50--150~word paragraph answers.}
  \label{tab:prompts-ideal}
  \footnotesize
  \begin{tabular}{llp{11.5cm}}
    \toprule
    \textbf{Variant} & \textbf{Strategy} \\
    \midrule
    1 & Simple: context + question, answer as JSON. \\
    2 & Biomedical expert persona, chain-of-thought before output. \\
    3 & Same as 2, explicit plain-text-only constraint (no lists or markdown). \\
    4 & Same as 3, explicit concise format instruction. \\
    5 & Same as 4, includes a few-shot example (P85-Ab biomarker). \\
    6 & XML-structured context and question tags, chain-of-thought embedded in JSON. \\
    7 & BioASQ eval-optimized: emphasizes recall, precision, no repetition, readability; 200-word single-paragraph. \\
    \bottomrule
  \end{tabular}
\end{table}
\subsection{Typed Generation Prompts}
Typed generation prompts (\texttt{prompts\_typed.json}) combine exact and ideal answer generation in a single call. Each variant produces both an \texttt{ideal\_answer} (paragraph summary) and an \texttt{exact\_answer} (typed JSON). Table~\ref{tab:prompts-typed} summarizes the variants.
\begin{table}[ht]
  \centering
  \caption{Typed generation prompts: combined exact + ideal answer variants by question type.}
  \label{tab:prompts-typed}
  \footnotesize
  \begin{tabular}{llp{11.5cm}}
    \toprule
    \textbf{Type} & \textbf{Variant} & \textbf{Strategy} \\
    \midrule
    \multirow{7}{*}{yes/no} & 1 & Chain-of-thought, outputs both ideal and exact. \\
    & 2 & Few-shot example (BRCA1), outputs both. \\
    & 3 & Two-sided evidence weighing, targets macro F1 by countering yes-bias. \\
    & 4 & Targets BioASQ manual evaluation criteria: recall, precision, no repetition, readability. \\
    & 5 & Dual few-shot examples (one yes, one no) to anchor both answer classes equally. \\
    & 6 & Mandatory evidence-quoting step before deciding. \\
    & 7 & Four real BioASQ few-shot examples: clear yes, indirect yes, clear no, tricky no. \\
    \midrule
    factoid & 1--2 & Chain-of-thought and few-shot variants; both output up to 5 ranked entities. \\
    \midrule
    list    & 1--2 & Chain-of-thought and few-shot variants; both output nested list of entities. \\
    \midrule
    summary & 1--2 & Chain-of-thought and few-shot variants; ideal answer only, 50--200 words. \\
    \bottomrule
  \end{tabular}
\end{table}
\subsection{Agent Quorum Prompts}
The agent quorum debate uses a structured multi-part prompt (Table~\ref{tab:prompts-quorum}) built from the components in \texttt{quorum/prompts.py}. Each agent turn receives: a system prompt establishing its role, a document sample with adaptive retention instructions, conversation history from prior rounds, and a thinking-focus description drawn from the focus pool.
\begin{table}[ht]
  \centering
  \caption{Agent quorum prompt components. The debate turn prompt is assembled from these sections; the final answer prompt uses a simpler synthesis-oriented structure.}
  \label{tab:prompts-quorum}
  \footnotesize
  \begin{tabularx}{\textwidth}{lX}
    \toprule
    \textbf{Component} & \textbf{Content} \\
    \midrule
    System prompt & ``You are a rigorous biomedical scientist participating in a structured expert debate. Your goal is to help reach the most accurate, evidence-grounded answer to a biomedical question. You respond exclusively with valid JSON.'' \\
    \midrule
    Document sample & Numbered documents with stable IDs, attached snippets, and adaptive retention rules (kept documents guaranteed in next round). \\
    \midrule
    Conversation history & Prior turns shown in full for recent rounds, compressed to one-line summaries for older rounds to stay within context-window bounds. \\
    \midrule
    Thinking focus & One of six cognitive lenses: analytical (logical decomposition), evidence-based (strict textual grounding), skeptical (gap-seeking), integrative (cross-source synthesis), pragmatic (precision-focused), or theoretical (mechanism-oriented). \\
    \midrule
    Agreement scale & \texttt{strongly\_disagree}, \texttt{disagree}, \texttt{agree}, \texttt{strongly\_agree} --- agents warned against premature strong agreement. \\
    \midrule
    Output format & \texttt{\{"opinion": "<reasoning>", "kept\_documents": [ids], "agreement": "<level>"\}}. \\
    \midrule
    Final synthesis & Receives all documents and a structured debate summary; produces a definitive answer with type-specific format. Must not reference the debate process. \\
    \bottomrule
  \end{tabularx}
\end{table}

\end{document}